\documentclass[lettersize,journal]{IEEEtran}
\usepackage{amsmath,amsfonts}
\usepackage{algorithmic}
\usepackage{array}
\usepackage[caption=false,font=normalsize,labelfont=sf,textfont=sf]{subfig}
\usepackage{textcomp}
\usepackage{stfloats}
\usepackage{verbatim}
\usepackage{graphicx}
\def\BibTeX{{\rm B\kern-.05em{\sc i\kern-.025em b}\kern-.08em
    T\kern-.1667em\lower.7ex\hbox{E}\kern-.125emX}}
\usepackage{balance}
\usepackage{times}
\usepackage{soul}
\usepackage{url}
\usepackage[utf8]{inputenc}
\usepackage{graphicx}
\usepackage{amsmath}
\usepackage{amsthm}
\usepackage{booktabs}
\usepackage{algorithm}
\usepackage{algorithmic}
\usepackage[switch]{lineno}
\usepackage{tikz}
\usepackage{tikz-qtree}
\usepackage{Figures/tikzstyle}

\usepackage{float} 
\usepackage{tabularx} 

\usepackage{bm}
\usepackage{amssymb}
\usepackage{amsmath}
\usepackage{bbm}
\usepackage{multirow}
\usepackage{lipsum}

\usepackage{pifont}
\newcommand{\cmark}{\ding{51}} 
\newcommand{\xmark}{\ding{55}} 

\usepackage{colortbl}
\usepackage{xcolor}
\usepackage{threeparttable} 

\usepackage{graphicx}  
\usepackage{wrapfig} 
\usepackage{xcolor}
\usepackage[
    colorlinks=true,       
    linkcolor=blue,       
    citecolor=black,      
    urlcolor=blue,      
    bookmarks=true,       
]{hyperref}

\definecolor{lightcoral}{rgb}{0.94, 0.5, 0.5}
\definecolor{lightgreen}{rgb}{0.56, 0.93, 0.56}
\definecolor{harvestgold}{rgb}{0.85, 0.57, 0.0}
\definecolor{brightlavender}{rgb}{0.75, 0.58, 0.89}
\definecolor{capri}{rgb}{0.0, 0.75, 1.0}
\definecolor{carminepink}{rgb}{0.92, 0.3, 0.26}
\definecolor{celadon}{rgb}{0.67, 0.88, 0.69}
\definecolor{darkpastelgreen}{rgb}{0.01, 0.75, 0.24}
\definecolor{darkgreen}{RGB}{150, 194, 145}
\definecolor{lightblue}{RGB}{138,160,205}
\definecolor{dark-blue}{RGB}{70,101,164}
\definecolor{brightlavender}{RGB}{186, 148, 209}
\definecolor{lightcoral}{RGB}{230, 164, 180}
\definecolor{darkpastelgreen}{RGB}{150, 194, 145}
\definecolor{harvestgold}{rgb}{0.85, 0.57, 0.0}
\begin{document}

\title{Empowering Time Series Analysis with Foundation Models: A Comprehensive Survey}


\author{Jiexia~Ye, Yongzi~Yu, Weiqi~Zhang, Le~Wang,  Jia~Li*, Fugee~Tsung \\
\thanks{*Corresponding Author: Jia Li}
\thanks{
Jiexia Ye and Yongzi Yu are with The Hong Kong University of Science and Technology (Guangzhou), Guangzhou, China (E-mail: \{jye324, yyu322\}@connect.hkust-gz.edu.cn). \\
Weiqi Zhang, Jia Li, and Fugee Tsung are with The Hong Kong University of Science and Technology, Hong Kong SAR, China (E-mail: wzhangcd@connect.ust.hk, \{jialee, season\}@ust.hk). Le Wang is with Shanghai University of Finance and Economics (E-mail: wangle2000@stu.sufe.edu.cn). This work has been submitted to the IEEE for possible publication. Copyright may be transferred without notice, after which this version may no longer be accessible.
}
}

\maketitle

\begin{abstract}
Time series data are ubiquitous across diverse real-world applications, making time series analysis critically important. Traditional approaches are largely task-specific, offering limited functionality and poor transferability. In recent years, foundation models have revolutionized NLP and CV with their remarkable cross-task transferability, zero-/few-shot learning capabilities, and multimodal integration capacity. This success has motivated increasing efforts to explore foundation models for addressing time series modeling challenges.
Although some tutorials and surveys were published in the early stages of this field, the rapid pace of recent developments necessitates a more comprehensive and in-depth synthesis to cover the latest advances. Our survey aims to fill this gap by introducing a modality-aware, challenge-oriented perspective, which reveals how foundation models pre-trained on different modalities face distinct hurdles when adapted to time series tasks. Building on this perspective, we propose a taxonomy of existing works organized by pre-training modality (time series, language, and vision), analyze modality-specific challenges and categorize corresponding solutions, discussing their advantages and limitations. Beyond this, we review real-world applications to illustrate domain-specific advancements, provide open-source code, and conclude with potential future research directions in this rapidly evolving field.

\end{abstract}

\section{Introduction}
Time series data, referring to sequences of observations collected over time, has been studied for centuries~\cite{SM-Mark1989}. Classical statistical models played a central role when data was relatively small and simple, such as ARIMA~\cite{kontopoulou2023review}, ARCH~\cite{SM-ARCH1982}, and Markov chains~\cite{al2023unique}. However, the rise of large-scale industrial systems in domains like transportation~\cite{Transur-2018}, IoT~\cite{IOTsur-2018}, and e-commerce~\cite{SM-Ecommerce2016} has generated vast, heterogeneous, and multimodal time series. These data exhibit complex and dynamic patterns that traditional statistical methods struggle to capture due to their reliance on pre-defined assumptions. 

Over the past few decades, deep learning has achieved remarkable success across domains such as computer vision (CV) and natural language processing (NLP). Unlike traditional statistical methods, these approaches can process large and diverse datasets in a more automated manner, reducing the need for manual feature engineering and enabling the discovery of complex patterns. Building on these successes, the time series community has progressively explored diverse architectures to address temporal modeling challenges~\cite{kong2025deep}.
Early efforts relied on recurrent neural networks (RNNs) for sequential memory, followed by convolutional neural networks (CNNs) for local feature extraction~\cite{TS-CNNC,TS-CNNF}. Graph neural networks (GNNs) were later introduced to capture relational dependencies~\cite{TSsur-graph2023,TS-graphF}, while transformers brought global attention mechanisms to the field~\cite{TSsur-tran2022}. The latest diffusion models have opened new directions for generative temporal representations~\cite{TSsur-dif2023}.

\begin{figure}[t]

  \centering
    \label{fig:fig_I_1}
  \includegraphics[width=0.49\textwidth]{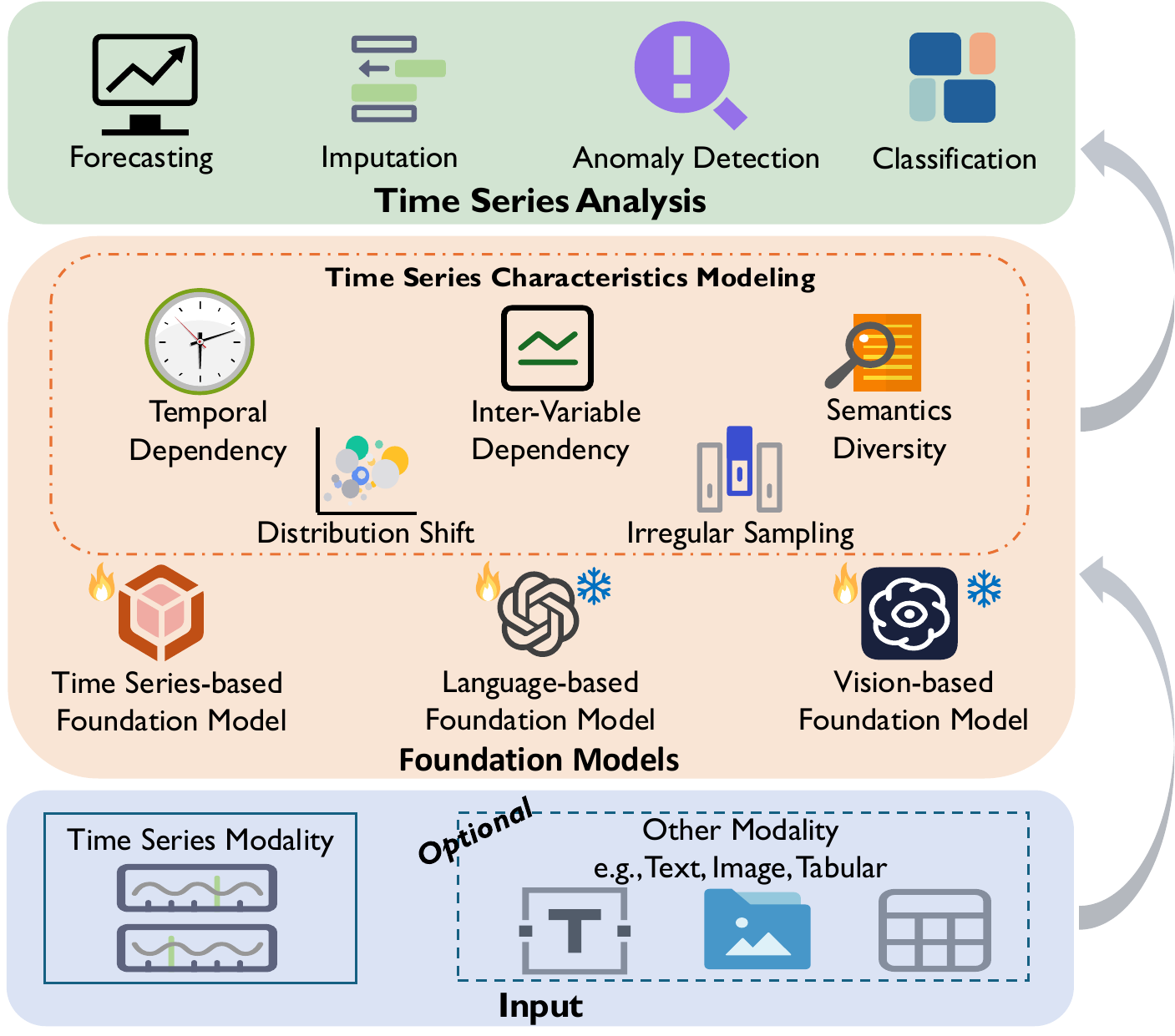}
  \vspace{-2mm}
  \caption{Foundation models for time series analysis can be developed through three approaches: pretraining foundation model from scratch on time series (time series based foundation model), or adapting language-based foundation models to time series tasks, or adapting vision-based foundation models to time series tasks.}
  \vspace{-5mm}
\end{figure}

However, these architectures remain predominantly task-specific and exhibit limitations when applied in broader contexts. First, they typically require substantial amounts of labeled data to achieve satisfactory performance, whereas many real-world scenarios—such as rare disease diagnosis~\cite{METS}—are characterized by severe data scarcity.
Second, their cross-domain transferability is fundamentally constrained: structural mismatches across datasets often makes models trained on one dataset incompatible with others, while even when transfer is feasible, distribution shifts in temporal patterns still lead to significant performance degradation~\cite{kim2021reversible}.
Third, most models operate in a unimodal setting and thus fail to leverage complementary modalities (e.g., text, images, graphs) —for instance, financial news that complement stock time series data~\cite{sawhney2021fast}.
Consequently, despite substantial advances, these approaches fall short of meeting the growing demand for unified frameworks with robust adaptability and generalization.

In recent years, foundation models (FMs) have achieved significant success in  NLP.  They are a general class of models pre-trained on large and diverse data and later adapted to a wide range of downstream tasks (e.g., via fine-tuning)~\cite{bommasani2021opportunities}. Early language-based foundation models (LFMs), such as BERT~\cite{BERT} and T5~\cite{T5}, established that large-scale pre-training can produce generic representations that substantially improve downstream performance. Subsequently, researchers further scaled up model and data sizes, giving rise to large language models (LLMs)—including GPT-2~\cite{GPT-2}, PaLM~\cite{PaLM}, and LLaMA~\cite{LLaMA}—which moved beyond representation learning and exhibited emergent abilities ~\cite{wei2022emergent}, such as in-context learning~\cite{GPT-3}, instruction following~\cite{hendrycks2020measuring}, and chain-of-thought reasoning~\cite{cot-2022}. These capabilities greatly enhance LFMs’ adaptability: they make few-shot and zero-shot learning more effective, allow diverse tasks to be unified through prompts and instructions, and support multimodal information integration. Since LLMs constitute the dominant class of LFMs, we use the two terms interchangeably in this survey. The capability of LLMs for broad generalization across tasks and domains has motivated the time series community to move beyond task-specific models toward general-purpose foundation models.

A dominant line of research focuses on adapting language-based foundation models (LFMs) for time series analysis, such as FPT~\cite{FPT}, Time-LLM~\cite{Time-LLM}, and TEMPO~\cite{TEMPO}. Under this paradigm, time series tasks are either reformulated to activate the reasoning capacity of LFMs, or LFMs are modified or augmented with modules specifically designed for temporal modeling; both approaches aim to harness the generalizable pre-trained knowledge of LFMs. The key challenges in adapting LFMs to time series analysis include devising effective encoding schemes to transform time series into representations that LFM can perceive, aligning temporal and textual modalities to enable cross-modal reasoning, and designing tuning strategies that balance accuracy with efficiency.
Despite recent progress, LFMs face intrinsic limitations for time series, since textual data are discrete and semantically grounded, whereas time series are continuous signals with dense dependencies and implicit semantics. These discrepancies limit their ability to capture complex temporal dynamics.

These limitations have motivated another research line: time series-based foundation models (TSFMs), which are pre-trained from scratch on time series data, as exemplified by Moirai~\cite{Moirai}, Lag-Llama~\cite{Lag-Llama}, and Moment~\cite{Moment}. This line avoids text-centric biases and the modality gap. But it still faces several major obstacles in building powerful TSFMs.
First, it requires overcoming the scarcity of large and diverse time series corpora, which hinders high-quality pre-training. 
Second, it demands effective encoding schemes and architectural designs that can accommodate diverse sequence structures while preserving generalizable temporal patterns in a scalable manner.
Finally, unifying heterogeneous time series tasks with varying granularity and context in a framework remains an open challenge.
Collectively, these obstacles underscore the complexity of building robust TSFMs.

\begin{table}[t]
\centering
\caption{Summary of our survey and relevant surveys}
\label{tab:tab_I_1}
\vspace{-2mm}
\resizebox{\linewidth}{!}{
\begin{threeparttable}
\begin{tabular}{lllllllll}
\midrule \midrule
\textbf{Survey} & \textbf{Year} & \textbf{Venue} & \textbf{Taxonomy} & \textbf{TSFM} & \textbf{LFM} & \textbf{VFM} & \textbf{\shortstack{Target \\ Papers}} & \textbf{Overlaps} \\
\midrule
Jin et al.\cite{zhou2023large}      & 2023 & arXiv & Data          & \cmark & \cmark & \xmark & 65 & 15 \\
Miller et al.\cite{Miller2024survey}   & 2024 & arXiv & Methodology   & \cmark & \cmark & \xmark & N/A & 11 \\
Jiang et al.\cite{jiang2024empowering} & 2024 & IJCAI & Pipeline    & \xmark & \cmark & \xmark & 21 & 17 \\
Zhang et al.\cite{zhang2024large}   & 2024 & IJCAI & Pipeline      & \xmark & \cmark & \xmark & 37 & 12 \\
Jin et al.\cite{jin2024position}    & 2024 & ICML  & Pipeline      & \xmark & \cmark & \xmark & 46 & 14 \\
Liang et al.\cite{liang2024foundation} & 2024 & KDD  & Methodology  & \cmark & \cmark & \xmark & 83 & 28 \\
kottapalli et al.\cite{kottapalli2025foundation} & 2025 & arXiv  & Multiple  & \cmark & \cmark & \xmark & 15 & 14 \\
Our paper                           & 2025 & --    & Challenge     & \cmark & \cmark & \cmark & 91 & 91 \\
\midrule \midrule
\end{tabular}

\begin{tablenotes}
\footnotesize
\item Note: Taxonomy denotes the main categorization employed in the survey. \textbf{TSFM} refers to Time Series-Based Foundation Model, \textbf{LFM} refers to Language-Based Foundation Model, and \textbf{VFM} refers to Vision-Based Foundation Model. We report the number of \textbf{Target Papers} investigated by these surveys and identify their \textbf{Overlaps} with our survey. The statistics are taken from their taxonomy tables or figures. \cite{Miller2024survey} focuses more on deep learning and does not provide an overall taxonomy table or figure.
\end{tablenotes}
\end{threeparttable}
}
\vspace{-4mm}
\end{table}

Beyond TSFMs, another emerging line of research seeks to overcome the limitations of LFMs by exploring vision-based foundation models (VFMs)— vision models pre-trained on massive image datasets. VFMs such as ViT~\cite{ViT}, MAE~\cite{MAE}, and BEiT~\cite{BEiT} have achieved remarkable success in vision, and motivated by the adaptation of LFMs to time series tasks, recent efforts have begun to investigate the potential of VFMs in this domain.
The rationale lies in the structural similarities between images and time series: both are real-world continuous signals with local correlations and multi-scale patterns, whereas language is discrete symbolic representation in the cognitive space. Applying VFMs to time series introduces challenges partly shared with LFMs—such as designing effective input representations and developing appropriate tuning strategies—but also distinct ones, arising from modality differences that require transforming temporal signals into image-like forms and adapting visual backbones to capture temporal dependencies.

Although differing in emphasis and challenges, these directions converge on the goal of building generalizable frameworks for time series analysis (as shown in Figure~\ref{fig:fig_I_1}). Recently, several surveys have begun to summarize the landscape as shown in Table~\ref{tab:tab_I_1}. Compared with these surveys, we place greater emphasis on the latest advancements and representative works in this field. Some surveys take a limited perspective, focusing mainly on LFM adaptations for time series~\cite{jiang2024empowering,zhang2024large,jin2024position}, while~\cite{Miller2024survey} concentrates on deep learning methods with only a section on foundation models. Unlike our survey, which centers on typical time series data,~\cite{zhou2023large} broadens its scope to general time series including video data, but only covers references up to 2023.~\cite{liang2024foundation} includes some models like PatchTST~\cite{PatchTST} and TimesNet~\cite{wu2022timesnet} while we consider these models as task-specific models. In addition, existing surveys primarily summarize papers from data-, pipeline-, or methodology-oriented perspectives. A largely underexplored aspect is the modality-aware challenge-oriented perspective, as FMs pre-trained on different modalities face distinct obstacles in time series adaptation. While existing works contribute by addressing these challenges, they have not been systematically organized in this way. To fill this gap, our survey categorizes the literature by pre-training modality—TSFMs, LFMs for time series, and VFMs for time series—and analyzes the unique challenges of each.
In doing so, our survey extends prior works with a broader and deeper perspective, and provides a timely and comprehensive overview of the latest advancements from challenge perspective in the field. Our main contributions are as follows:

(1) \textbf{Comprehensive and up-to-date survey.} We provide a comprehensive and up-to-date survey on foundation models for time series analysis, covering time series–based foundation models, language-based and vision-based foundation models adapted to time series tasks.

(2) \textbf{Novel Modality-Aware, Challenge-Oriented Taxonomy.} We propose a taxonomy that organizes existing works according to their pre-training modalities and systematically examines the inherent challenges of each category. Within this framework, we further categorize and analyze the corresponding solutions of each challenge, offering a structured perspective on how modality-specific factors influence the adaptation of foundation models to time series tasks.

(3) \textbf{Applications, resources, and future opportunities.} 
We provide code resources to facilitate access to existing implementations, review representative applications of foundation models across diverse time series domains, and identify promising research avenues to advance in this field.

\textit{Organization of Survey}:
Section~\ref{sec:background} introduces the background of foundation models and time series analysis. Section~\ref{sec:time} examines key challenges in pre-training TSFMs. Section~\ref{sec:text} addresses adaptation issues when applying LFMs to time series tasks, while Section~\ref{sec:image} focuses on similar challenges for adapting VFMs.
Section~\ref{sec:applications} investigates representative applications in diverse domains. Section~\ref{sec:future} explores promising research directions, and Section~\ref{sec:conclusion} concludes the paper.

\begin{figure}[ht]
  \centering
  \includegraphics[width=0.49\textwidth]{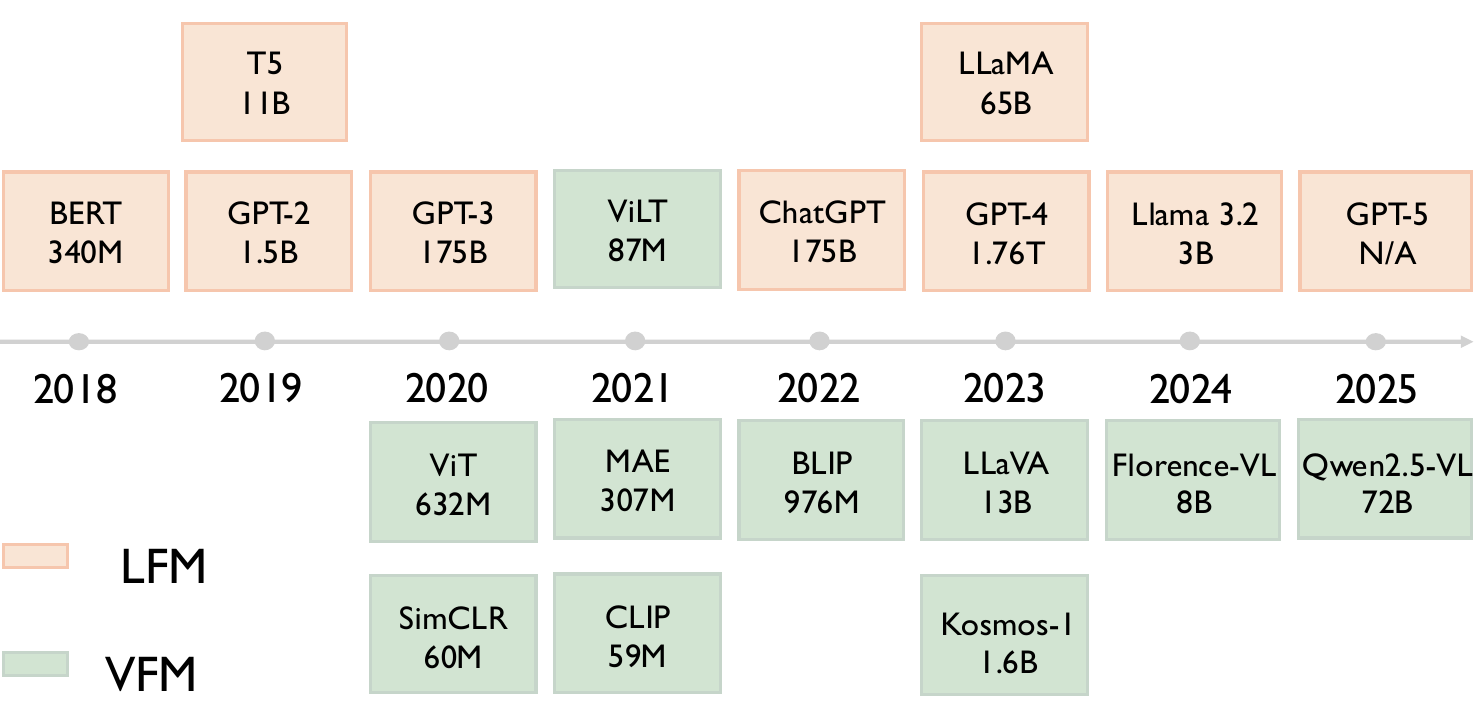}
    \vspace{-4mm}
  \caption{A roadmap of representative language-based foundation models (LFMs) and vision-based foundation models (VFMs) and their corresponding parameter scale. The reported values of parameter scale correspond to commonly adopted configurations in the literature, and may vary across implementations.
  \textbf{M} is million while \textbf{B} is billion.}
  \label{fig:fig_II_1}
    \vspace{-4mm}
\end{figure}

\section{Background}
\label{sec:background}
\subsection{Foundation Models} 
The term \textit{Foundation Model (FM)} was coined in~\cite{bommasani2021opportunities} to describe a broad category of models initially pre-trained on large and diverse datasets, subsequently adapted for various downstream tasks through methods such as fine-tuning. Unlike conventional task-specific models, they have been successfully developed in  NLP and CV and exhibit remarkable versatility, enabling adaptation across diverse tasks. In this context, we provide an overview of the development of FMs in NLP and CV to introduce background for their adaptation to time series analysis (as shown in Figure \ref{fig:fig_II_1}).

\textbf{Language-based Foundation Models}
Language-based Foundation Models (LFMs) are those FMs pre-trained on massive textual datasets (e.g., books, websites, and multilingual corpora) and they have made significant advances in NLP. As these models scale in both data volume and parameter size, they become large language models (LLMs)—a subset of foundation models. Key milestones in the development of LFMs include the introduction of BERT~\cite{BERT} in 2018, which pioneered transformer-based pre-training, laying groundwork for modern LLMs with context-aware language understanding; GPT-2~\cite{GPT-2} in 2019, which proved large decoder-only transformers excel at zero-shot learning and T5~\cite{T5} in the same year, which unified NLP tasks as text-to-text problems; GPT-3~\cite{GPT-3} in 2020  which, with its 175 billion parameters, achieved unprecedented performance in various tasks; ChatGPT in 2022  which highlighted the conversational applications of LLMs; LLaMA~\cite{LLaMA} in 2023 which democratized access to high-performance language models, fostering open-source innovation; GPT-4~\cite{GPT-4} in 2023, bringing enhanced reasoning and comprehension abilities; Llama 3.2 in 2024~\cite{dubey2024llama}, introducing multimodal “Vision” models that support both text and image inputs; GPT-5 in 2025, advancing reasoning and multimodal integration.

\textbf{Vision-based Foundation Models}
In this survey, we use the term vision-based foundation models (VFMs) broadly to include both large vision models and vision–language models, which are pre-trained on large-scale visual datasets or multimodal corpora containing paired images and text.
A key milestone was ViT in 2020~\cite{ViT}, which showed that transformers could outperform CNNs in image recognition when trained at scale. Building on contrastive self-supervision, SimCLR in 2020~\cite{SimCLR} demonstrated the power of large-scale label-free pre-training. MAE in 2022~\cite{MAE} advanced this paradigm with masked autoencoding for scalable and efficient visual representation learning. In the direction of vision-language models, CLIP in 2021~\cite{CLIP} aligned images and text at web scale, while ViLT in 2021~\cite{ViLT} streamlined fusion by directly embedding image patches. BLIP in 2022~\cite{BLIP} improved adaptability with flexible vision–language pre-training pipelines. LLaVA in 2023~\cite{LLaVA} integrated LLaMA with visual encoders for instruction-following. Kosmos-1 in 2023~\cite{Kosmos-1} pushed VFMs toward general-purpose multimodal assistants. Florence-VL in 2024~\cite{chen2025florence} introduced enriched multi-depth fusion techniques to enhance vision-language alignment. Qwen2.5-VL~\cite{bai2025qwen2} in 2025 pushed real-time multimodal comprehension.

LFMs and VFMs exhibit strong cross-task generalization from large-scale pre-training, suggesting similar benefits for time series analysis with limited labels and diverse distributions. Moreover, time series share structural similarities with both language and vision: like language, they are sequential with contextual dependencies, and like images, they are continuous-valued signals with local and multi-scale patterns. These similarities provide a natural basis for adapting LFMs and VFMs to time series tasks.

\begin{figure*}[ht]
  \centering
  \includegraphics[width=0.99\textwidth]{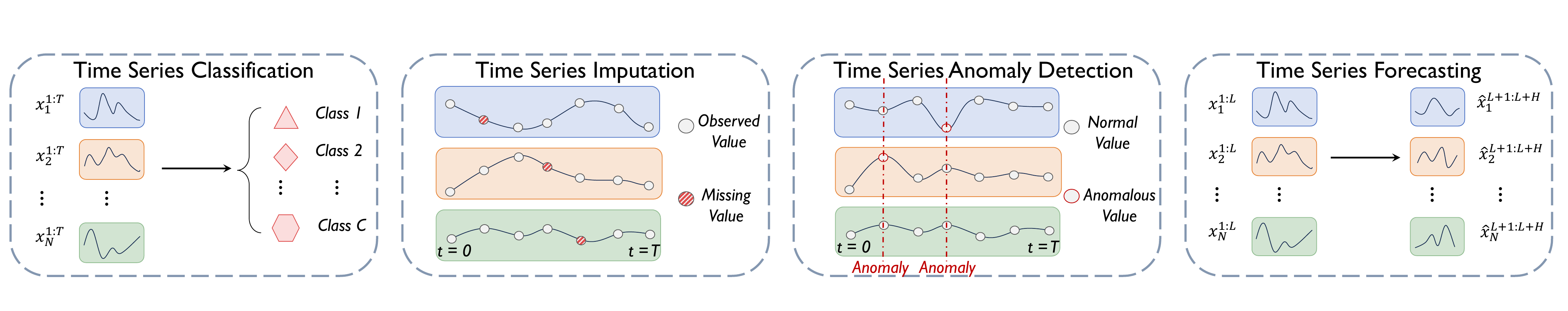}
      \vspace{-2mm}
  \caption{Illustration for typical time series tasks: Classification, Imputation, Anomaly Detection, and
Forecasting.}
  
  \label{fig:fig_II_2}
      \vspace{-4mm}
\end{figure*}

\subsection{Time Series} 
\textbf{Time Series} Time series are composed of data points collected at successive intervals, with sampling frequencies ranging from high (e.g., financial signals)~\cite{SSPT} to low (e.g., annual climate records)~\cite{liu2024combining}. In practice, sampling may not always be uniform, leading to irregularly spaced observations that further complicate analysis~\cite{DAM}. Time series data can be categorized as univariate or multivariate depending on the number of channels. A univariate time series of length $T$ is denoted as $\bm{X}=\left\{ x_t\right\}_{t\in\left\{1, ...T\right\}}\in \mathbb{R}^{T}$, where $x_t$ represents the observation at timestamp $t$. A multivariate time series with $N$ $(N>1)$ channels is expressed as $\bm{X}=\left\{\bm{x_t}\right\}_{t\in\left\{1, ...T\right\}}\in \mathbb{R}^{N\times T}$, where each $\bm{x}_t$ is an $N$-dimensional vector recording signals from all channels at time $t$.

\textbf{Channel-Independence vs. Channel-Mixing} 
In this survey, we categorize channel handling in time series models into two settings. 
In the \textit{channel-independence} setting, each channel is treated as an isolated univariate sequence, processed separately without explicit mechanisms for inter-channel communication. Any integration is deferred to a late stage, such as feature concatenation or averaging, implying that temporal dependencies dominate while cross-channel correlations are either negligible or only implicitly captured. This setting naturally applies to univariate datasets and can also be adopted for  multivariate datasets when channels are modeled independently~\cite{TEMPO,FPT}. 
In contrast, the \textit{channel-mixing} setting explicitly models interactions across channels through mechanisms such as cross-channel attention~\cite{GTT, TTM}. This paradigm inherently applies to multivariate datasets, assuming that inter-channel relationships provide essential complementary information for downstream tasks.

\subsection{Time Series Tasks} 
Time series analysis primarily focuses on four mainstream tasks: forecasting (predicting future values from historical observations), classification (assigning labels to entire sequences or specific time points), anomaly detection (identifying irregular or abnormal patterns), and imputation (estimating missing values). Beyond these well-established tasks (as illustrated in Fig.~\ref{fig:fig_II_2}), emerging directions such as event prediction~\cite{TimeCAP} and time-series question answering~\cite{ChatTime} are further broadening the scope of time series analysis.

\textbf{Time Series Classification}
 Given a multivariate time series data $\bm{X}$, the time series classification model $f:\mathbb{R}^{N\times T} \rightarrow \mathbb{R}^{C}$ aims to distinguish the time series data of $C$ different categories by 
 \begin{equation*}
     I^c = f(\bm{X}), 
 \end{equation*}
 and return the class index $I^c$.

\textbf{Time Series Imputation}
The goal of time series imputation is to recover the value of missing observations precisely. To describe the data missingness of time series $\bm{X}$,  we denote $\bm M\in \{0,1\}^{N\times T}$ as the mask matrix, where $\bm X_{i,j}$ can be observed only if $\bm M_{i,j}=1$.
Given the observed time series data $\bm X^{\star}$, mask matrix $\bm M$, the imputation algorithm aims to recover the missingness of data matrix by
\begin{equation*}
    \bm{\hat{X}}=f\left(\bm X^{\star}\right),
\end{equation*}
where  $\bm X^{\star}_{i,j}= \begin{cases}\mathrm{NA}, & \text{if } \bm M_{i,j}=0 \\ \bm X_{i,j}, & \text { otherwise }\end{cases}$, $\mathrm{NA}$ represents a missing value, and $f:(\mathbb{R} \cup\{\mathrm{NA}\})^{N\times T} \rightarrow \mathbb{R}^{N\times T}$ denotes a learnable imputation function.

\textbf{Time Series Anomaly Detection} At timestamp $t$, the anomaly detection model is developed to find out whether there exist anomalies in the past time series subsequence. Given a multivariate time series subsequence $\left\{\bm{x_{t-w+1}}, ..., \bm{x_{t}}\right\}$ with window size $w$, the multivariate time series anomaly detection model $f: \mathbb{R}^{N\times W}\rightarrow \left\{0,1\right\}$ aims to return an anomaly indicator $I^a_t\in \left\{0,1\right\}$ by 
\begin{equation*}
    I^a_t=f\left(\left\{\bm{x_{t-w+1}}, ..., \bm{x_{t}}\right\}\right),
\end{equation*}
where there is indeed an anomaly in subsequence if $I^a_t=1$, the subsequence has no anomaly otherwise.

\textbf{Time Series Forecasting}
Time series forecasting aims at predicting the future value with horizon $H$ based on the input time series with lookback window $L$. Given the input time series data $\bm{X}_L = \left\{ \bm{x_1},  ..., \bm{x_L}\right\}$, the forecasting model provide the predicted value for future $H$ timestamps by 
\begin{equation*}
    \left\{ \bm{\hat{x}_{L+1}},  ..., \bm{\hat{x}_{L+H}}\right\} = f (\bm{X}_L),
\end{equation*}
where $f:\mathbb{R}^{N\times L}\rightarrow \mathbb{R}^{N\times H}$ is a learnable forecasting function. Meanwhile, considering the length of the forecasting horizon, the task could be further divided into long-term/short-term time series forecasting. Based on the quantity of training data, few-shot/zero-shot forecasting pipeline could also be established.

\subsection{Time Series Characteristics} 
Unlike text and image data, time series exhibits distinctive characteristics, including temporal dependency, inter-variable dependency, distribution shifts, irregular sampling, and semantic diversity. These properties reflect the intrinsic complexity of time series and impose unique challenges on developing foundation models that are robust and generalizable for time series analysis.

\textbf{Temporal Dependency}
Time series data, recorded at successive timestamps, inherently exhibit temporal dependencies where past observations influence future values. These dependencies occur at multiple scales: short-term dependency captures local dynamics; long-term dependency reflects trends or recurring patterns; and seasonality represents a special form of long-term dependency characterized by predictable cycles such as daily, weekly, or yearly patterns~\cite{TEMPO}. These multi-scale patterns require foundation models for time series analysis to effectively capture dependencies at different granularities while preserving the inherent sequential order and causal relationships.

\textbf{Inter-variable Dependency} 
In multivariate time series, each variable (or channel) represents a distinct aspect of a complex system, and their interactions convey essential information about its dynamics. These inter-variable dependencies allow information from one variable to complement others, and modeling them jointly can reveal latent patterns that individual variables alone cannot capture. In cross-dataset scenarios, not only may datasets differ in number of channels, but the nature and strength of inter-channel collaborations can also vary significantly, posing challenges for the design of foundation models that aim to generalize across diverse domains~\cite{UNITS,GTT}.

\textbf{Distribution Shift}
In many real-world scenarios, the distribution of time series data changes across time or domains. Such shifts can occur in the marginal distribution $P(X_t)$, reflecting changes in value ranges or statistical moments, or in the conditional distribution $P(X_t \mid X_{t-1}, \dots)$, indicating changes in dependency structures. Distribution shift describes discrepancies between training and test distributions, which significantly challenge model generalization~\cite{Kim2022ReversibleIN}, especially in cross-domain or cross-dataset settings. This underscores the need for robust foundation models to mitigate distribution shifts and ensure reliable, accurate predictions across diverse datasets.

\textbf{Irregular Sampling} In some real-world scenarios, time series data are collected at uneven or event-driven intervals due to the nature of the application, such as clinical measurements scheduled at varying times, trades occurring irregularly in financial markets, or condition-triggered recordings in industrial systems. Such irregularity disrupts the continuity of temporal patterns and complicates temporal alignment. For foundation models, it introduces challenges in learning consistent temporal representations, managing variable time gaps, and integrating information across heterogeneous sampling rates without losing essential dynamic characteristics~\cite{DAM}.

\textbf{Semantics Diversity} Unlike image and text data, where consistent semantics can often be found across different domains (with each word or visual patch representing similar meanings in various sentences or images), time series data lacks this uniformity. Identical subsequences or shapelets in time series datasets can represent entirely different concepts depending on the context. This semantic variability—especially across domains—complicates representation learning and challenges the transferability of foundation models~\cite{Unisur-2024}.

\begin{figure*}[ht]
  \centering
  \includegraphics[width=0.99\textwidth]{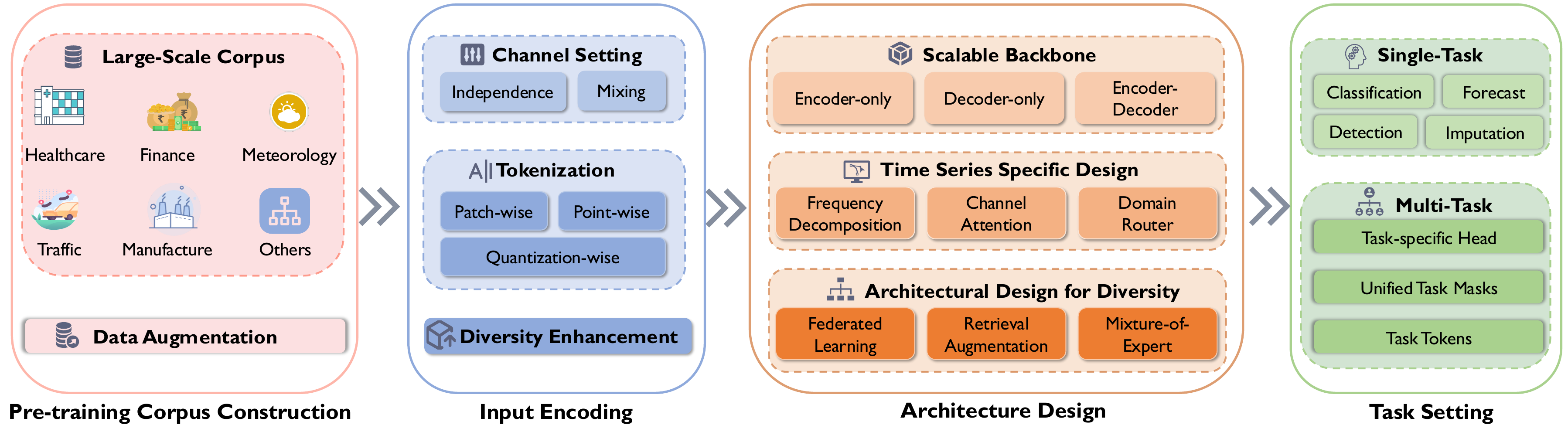}
      \vspace{-2mm}
  \caption{We identify four key challenges for building robust time series-based foundation models and summarize existing solutions to each dimension. }
      \vspace{-4mm}
  \label{fig:fig_III_1}
\end{figure*}

\begin{table}[!ht]
\centering
\caption{Summary of Large-scale Time Series Corpora}
\begin{tabular}{llll}
\midrule \midrule
\textbf{Work} & \textbf{Corpus} & \textbf{Time Points} & \textbf{Domains} \\
\midrule
Moment \cite{Moment} & Time Series Pile & 1.23 billion & 13 \\
Moirai \cite{Moirai} & LOTSA & 27 billion & 9 \\
Timer \cite{Timer} & UTSD & 1 billion & N/A \\
Time-MoE \cite{Time-MoE} & Time-300B & 300 billion & N/A \\
Sundial \cite{Sundial} & TimeBench & 1 trillion & N/A \\
\midrule \midrule
\end{tabular}
\label{tab:tab_III_1}
\end{table}

\section{Time Series-based Foundation Model}
\label{sec:time}
\begin{table*}[!ht]
\caption{A summary of Time Series-Based Foundation Models}

\resizebox{\linewidth}{!}{
\begin{threeparttable}
\begin{tabular}{@{}llllllll@{}}
\midrule \midrule

\textbf{Model} &
  \textbf{Architecture} &
  \textbf{Tokenization} &
  \textbf{\begin{tabular}[c]{@{}l@{}}Channel\\  Setting\end{tabular}} &
  \textbf{\begin{tabular}[c]{@{}l@{}}Model Size\\  (Max)\end{tabular}} &
  \textbf{\begin{tabular}[c]{@{}l@{}}Pre-training \\  Data Scale\end{tabular}} &
  \textbf{Tasks} &
  \textbf{Code} \\ \midrule
TimeDiT\cite{TimeDiT} &
  Encoder-only &
  Point-wise &
  Mixing &
  N/A &
  N/A &
  F(r)/I/D &
  N/A \\
DAM\cite{DAM} &
  Encoder-only &
  Point-wise &
  Independence &
  N/A &
  44M (s) &
  F(o) &
  N/A \\
ForecastPFN\cite{ForecastPFN} &
  Encoder-only &
  Point-wise &
  Independence &
  5.72M &
  N/A &
  F(o) &
  \href{https://github.com/abacusai/forecastpfn}{Link} \\
TimeCLR\cite{TimeCLR} &
  Encoder-only &
  Point-wise &
  Independence &
  0.71M &
  N/A &
  C &
  \href{https://github.com/Josu16/TimeCLR-Michael}{Link} \\
Moirai\cite{Moirai} &
  Encoder-only &
  Patch-wise &
  Mixing &
  311M &
  27B (t) &
  F(r) &
  \href{https://github.com/redoules/moirai}{Link} \\
UNITS\cite{UNITS} &
  Encoder-only &
  Patch-wise &
  Mixing &
  30.77M &
  35M (t) &
  F(o)/I/D/C &
  \href{https://github.com/mims-harvard/UniTS}{Link} \\
GTT\cite{GTT} &
  Encoder-only &
  Patch-wise &
  Mixing &
  57M &
  2.4B (t) &
  F(o) &
  \href{https://github.com/cfeng783/GTT}{Link} \\
Moment\cite{Moment} &
  Encoder-only &
  Patch-wise &
  Independence &
  385M &
  1.23B (t) &
  F(o)/I/D/C &
  \href{https://huggingface.co/AutonLab}{Link} \\
FFTS\cite{FFTS} &
  Encoder-only &
  Patch-wise &
  Independence &
  N/A &
  N/A &
  F(o)/I/D/C &
  \href{https://github.com/shengchaochen82/FFTS}{Link} \\
LPTM\cite{LPTM} &
  Encoder-only &
  Patch-wise &
  Independence &
  100M &
  N/A &
  F(o)/C &
  \href{https://github.com/AdityaLab/Samay}{Link} \\
Chronos\cite{Chronos} &
  Encoder-Decoder &
  Quantization-wise &
  Independence &
  710M &
  84B (t) &
  F(r) &
  \href{https://github.com/amazon-science/chronos-forecasting}{Link} \\
TOTEM\cite{TOTEM} &
  Encoder-Decoder &
  Quantization-wise &
  Independence &
  N/A &
  N/A &
  F(o)/I/D &
  \href{https://github.com/SaberaTalukder/TOTEM}{Link} \\
TimeGPT\cite{TimeGPT} &
  Encoder-Decoder &
  Patch-wise &
  Independence &
  N/A &
  100B (t) &
  F(r)/D &
  N/A \\
TTM\cite{TTM} &
  Encoder-Decoder &
  Patch-wise &
  Independence &
  1M &
  1B (s) &
  F(o) &
  \href{https://github.com/ibm-granite/granite-tsfm/tree/main/tsfm\_public/models/tinytimemixer}{Link} \\
TimeRAF\cite{TimeRAF} &
  Encoder-Decoder &
  Patch-wise &
  Independence &
  N/A &
  320M(t) &
  F(o) &
  N/A \\
ROSE\cite{ROSE} &
  Encoder-Decoder &
  Patch-wise &
  Independence &
  4.5M &
  887M(t) &
  F(o) &
  \href{https: //github.com/decisionintelligence/ROSE}{Link} \\
Lag-Llama\cite{Lag-Llama} &
  Decoder-only &
  Point-wise &
  Independence &
  200M &
  352M (t) &
  F(r) &
  \href{https://github.com/time-series-foundation-models/lag-llama}{Link} \\
Time-MoE\cite{Time-MoE} &
  Decoder-only &
  Point-wise &
  Independence &
  2.4B &
  300B (t) &
  F(o) &
  \href{https://github.com/Time-MoE/Time-MoE}{Link} \\
MOIRAI-MOE\cite{MOIRAI-MOE} &
  Decoder-only &
  Patch-wise &
  Mixing &
  935M &
  N/A &
  F(ro) &
  N/A \\
Sundial\cite{Sundial} &
  Decoder-only &
  Patch-wise &
  Independence &
  444M &
  1T &
  F(r) &
  \href{https://github.com/thuml/Sundial}{Link} \\
Timer\cite{Timer} &
  Decoder-only &
  Patch-wise &
  Independence &
  67M &
  28B (t) &
  F(o)/I/D &
  \href{https://github.com/thuml/Large-Time-Series-Model}{Link} \\
TimesFM\cite{TimesFM} &
  Decoder-only &
  Patch-wise &
  Independence &
  200M &
  100B (t) &
  F(o) &
  \href{https://github.com/google-research/timesfm}{Link} \\
TimesFM-ICF\cite{TimesFM-ICF} &
  Decoder-only &
  Patch-wise &
  Independence &
  N/A &
  400B &
  F(o) &
  N/A

  \\ \midrule \midrule
  
\end{tabular}

\begin{tablenotes}
\footnotesize
\item Note: \textbf{M} is million while \textbf{B} is billion. \textbf{s} denotes samples while \textbf{t} denotes time points. Task setting: \textbf{F(o)} is Point  Forecast; \textbf{F(r)} is Probabilistic Forecast; \textbf{I} is Imputation; \textbf{D} is Anomaly Detection; \textbf{C} is Classification. \textbf{N/A} denotes \textit{Not Applicable}.
\end{tablenotes}
\end{threeparttable}

}
\label{tab:tab_III_2}
\end{table*}
This section focuses on \textit{Time Series-based Foundation Model (TSFM)}—foundation model pre-trained entirely on time series data from scratch. Their goal is to enable effective transfer learning across diverse time series downstream  tasks. Four critical dimensions shape their transferability: first, the scale and diversity of the pre-training corpus which determines the breadth and depth of temporal knowledge acquired; second, the time series encoding techniques to accommodate diverse data types, model time series characteristics to enhance model comprehension; third, the architectural design, which supports parameter scalability and time series-aware design, enabling robust capture of diverse temporal patterns; finally, the task configuration that ensures compatibility with various task types. By examining these challenges (as shown in Figure~\ref{fig:fig_III_1}), we aim to summarize current methodologies and highlight ongoing efforts to address them.

\subsection{Pre-training Corpus Construction}
Pre-training foundation models for time series from scratch critically depends on the availability of large-scale, high-quality datasets. Chronos~\cite{Chronos} highlights that the quantity and quality of pre-training data often outweigh architectural innovations in determining performance. However, compared to the massive and diverse corpora in NLP and CV, publicly available time series datasets (e.g., Monash~\cite{TSdat-Monash}, UCI~\cite{TSdat-UCI}) are significantly smaller, fragmented. This scarcity constitutes a fundamental bottleneck for time series based foundation models, severely limiting their capacity to acquire broad and transferable temporal knowledge. There are two complementary directions to address this challenge: (1) constructing large-scale, diverse corpora through integration or synthetic generation, and (2) developing data augmentation strategies to enrich limited datasets. Both approaches can substantially enhance pre-training robustness, while TSFMs that do not adopt them often compensate for limited data through advanced time series encoding or architectural innovations.

\textbf{Large-scale Corpus}
Several efforts focus on collecting and integrating massive, diverse time series corpora in Table~\ref{tab:tab_III_1}. For example, Moment~\cite{Moment} constructs the Time Series Pile, integrating 1.23 billion timestamps from 13 domains. Moirai~\cite{Moirai} introduces LOTSA, an open dataset collection with 27 billion observations across 9 domains for forecasting. Timer~\cite{Timer} curates the Unified Time Series Dataset (UTSD) with 1 billion time points, establishing quality criteria and a hierarchical structure to support large-scale pre-training of TSFM. Time-MoE~\cite{Time-MoE} develops a data-cleaning pipeline and creates Time-300B, the largest collection to date, with over 300 billion time points for foundation model pre-training. Sundial~\cite{Sundial} collects and curates TimeBench, exceeding one trillion time points from multiple sources, primarily real-world records with 0.05\% synthetic data added to increase pattern diversity. These efforts demonstrate the feasibility of large-scale corpus construction. However, they are costly, heavily reliant on real-world records, and still constrained by limited domain balance, data quality and pattern diversity.

\textbf{Data Augmentation} Beyond dataset construction, data augmentation has emerged as a key technique to expand the scale of time series data and diversify its distributions while preserving temporal integrity.
Lag-Llama~\cite{Lag-Llama} applies Freq-Mix and Freq-Mask to generate additional training samples to mitigate overfitting. TimeCLR~\cite{TimeCLR} leverages augmentations to strengthen robustness against warping and noise. Chronos~\cite{Chronos} enhances pattern diversity through Mixup and expands datasets using KernelSynth-generated synthetic series. Distinct from these approaches, ForecastFN~\cite{ForecastPFN} highlights the limitations of pre-training on limited real-world data and instead constructs a synthetic corpus with diverse, general distributions to enable more consistent zero-shot forecasting performance.
In summary, data augmentation plays a vital role in enriching temporal patterns, improving robustness to distortions and noise, and mitigating data scarcity.

\subsection{Time Series Input Encoding}
Encoding raw time series into representations suitable for foundation models is non-trivial, as time series data exhibit unique structural challenges compared to text or images. First, the multivariate nature of time series challenges pre-training, as datasets often contain heterogeneous channel numbers which requires flexible encoding schemes. Second, designing tokenization strategies that preserve temporal dependencies while remaining compatible with transformer architectures is critical for capturing intrinsic sequential patterns.
Third, the diversity of pre-training scenarios requires encoding schemes that can flexibly handle heterogeneous conditions to enhance model generalization.

\textbf{Channel Setting} 
Pre-training corpora for time series typically include both univariate and multivariate datasets. 
For univariate time series, usually, each univariate sequence undergoes Reversible Instance Normalization (RevIN)~\cite{RevIN} to mitigate distribution shifts, followed immediately by tokenization to generate the model input tokens.
For multivariate time series, pre-training requires careful consideration of how channels are organized and represented. Most existing works adopt the channel-independence setting, where channels are treated as separate univariate sequences processed by a shared model independently~\cite{UniTime}. This strategy simplifies the input structure, alleviates cross-variable complexity, and enables consistent pre-training across datasets with heterogeneous channel numbers.
In contrast, channel-mixing setting captures multi-channel interactions~\cite{UniTime} among different channels. Most methods under channel-mixing setting retain the original channel structure at the input level (sometimes with patching or flattening~\cite{MOIRAI-MOE,Moirai} for standardization), while deferring the cross-channel correlations modeling to later model design. TimeDiT~\cite{TimeDiT} is a notable exception that enforces a fixed channel dimensionality by padding or segmenting inputs.

\textbf{Time Series-Aware Tokenization} We summarize three approaches for time series tokenization in TSFMs: patch-wise, point-wise and quantization-wise tokenization. Beyond merely formatting inputs, effective tokenization is expected to retain or extract essential temporal characteristics of the data. 
\textit{Patch-wise tokenization} is predominant. It segments time series into contiguous patches, enabling the preservation of local semantics, supporting longer input sequences, and reducing computational costs.
\textit{Point-wise tokenization} treats each individual time step as a token, which allows for fine-grained temporal modeling.
Representative approaches include Lag-Llama~\cite{Lag-Llama}, which constructs tokens using lagged features and date-time information to capture frequency and periodicity, and Time-MoE~\cite{Time-MoE}, which directly embeds each time point as a token to retain complete temporal information.
TimeDiT~\cite{TimeDiT} directly treats raw values at each time step as modeling units and applies masking in the original point space.
Both ForecastPFN~\cite{ForecastPFN} and DAM~\cite{DAM} represents each time-value pair as a token.
\textit{Quantization-wise tokenization} methods transform continuous time series into discrete token sequences, enabling foundation models to process temporal data in a manner similar to text.
Chronos~\cite{Chronos} adopts a scaling-and-quantization scheme that maps values into discrete bins, making them compatible with T5 backbone.
TOTEM~\cite{TOTEM} proposes Tokenized Time Series Embeddings, a tokenization method that learns a fixed codebook of discrete tokens to represent univariate waveform shapes, providing a generic and domain-agnostic encoding for time series.

\textbf{Time Series Diversity Enhancement}
To address the challenges foundation models face in handling diverse and complex time series data, recent encoding methods introduce diversity enhancements. For example, unlike fixed-length patching strategy, LPTM~\cite{LPTM} employs self-supervised pre-training to dynamically determine the optimal segmentation length for each dataset; DAM~\cite{DAM} expands applicability from regularly sampled data to irregular and variable-length data by using a history sampling regime, balancing global perspectives with attention to local temporal trends;   These innovations collectively improve the robustness and generalization of foundation models across heterogeneous time series domains.

\subsection{Architecture Design}
Architecture design in TSFMs aims to accommodate the diversity of pre-training corpus and capture diverse temporal patterns, while ensuring computational efficiency and enhancing generalization to downstream tasks.
It requires developing not only scalable backbones that support efficient large-scale training and transferability, but also time series–specific designs that account for the unique properties of temporal data. In addition, we summarize that some works introduce mechanisms beyond time series–specific considerations to equip TSFMs with broader functionality to improve efficiency and generalization from different dimensions.

\textbf{Scalable Backbone}
Transformer has become the dominant backbone for time series foundation models due to its scalability and parallel processing efficiency. While any scalable deep learning architecture could serve this purpose, Timer~\cite{Timer} extensively evaluates candidates and confirms transformer as the most effective backbone. 
Sundial~\cite{Sundial} further extends the scalability frontier of transformer by introducing a generative backbone equipped with efficient architectural adaptations such as RoPE, FlashAttention, and multi-patch prediction. Sundial also validates scaling laws for TSFMs.
An exception is TTM~\cite{TTM}, which combines MLP blocks with gated attention from TSMixer, creating a lightweight pre-trained backbone for efficient zero/few-shot forecasting. Transformer-based TSFMs typically follow encoder-only, decoder-only, or encoder-decoder architectures. Generally speaking, encoder-only models make predictions based on the full context, making them ideal for classification. Decoder-only models generate tokens autoregressively, favoring long-horizon forecasting.~\cite{yao2025towards} further shows that encoder-only models generally scale more favorably than decoder-only models on in-distribution data. Encoder-decoder models are the least used due to their high computational cost. 

\textbf{Time Series Specific Design}
Most TSFMs adopt transformer architectures without inductive biases, which may limit their ability to capture complex temporal patterns beyond long-range dependencies. To address this, several works integrate time-series-specific designs into transformer structure to model diverse time series characteristics. For multivariate dependency, 
TTM~\cite{TTM} adopts a channel-independent pre-training scheme but models cross-channel interaction by introducing the decoder’s channel-mixer block. GTT~\cite{GTT} integrates a dedicated channel-attention module into the self-attention block, allowing the model to explicitly aggregate inter-channel dependencies.
UNITS~\cite{UNITS} applies two-way self-attention across both the temporal and variable dimensions, thereby enabling it to simultaneously capture temporal dynamics and cross-variable interactions in a unified framework. 
Moirai~\cite{Moirai} employs multi-patch projections to handle diverse time series frequencies and proposes any-variate attention, treating all variates as a single sequence to address varying dimensionality. 
MOIRAI-MOE~\cite{MOIRAI-MOE} also flattens all channels into a single long sequence, so that causal attention can naturally span across variables.
For temporal dependency, 
Time-MoE~\cite{Time-MoE} introduces sparse temporal mixture-of-expert layers with sparse expert activation and domain-aware routing to efficiently model heterogeneous temporal patterns.
ROSE~\cite{ROSE} incorporates decomposed frequency learning, where fourier-based decomposition with frequency masking disentangles complex temporal patterns.
DAM~\cite{DAM} proposes token merging and cross-attention to efficiently integrate temporal and frequency features.
MOIRAI-MOE~\cite{MOIRAI-MOE} proposes a novel expert gating function to dynamically route each token to specialized experts within transformer layers, enabling the model to learn diverse temporal patterns.

\textbf{Architectural Design for Diversity}
While time series–specific designs focus on tailoring architectures to the intrinsic properties of temporal data, some work augments TSFMs with novel mechanisms to enhance adaptability to diverse time series datasets.
 FFTS~\cite{FFTS} introduces a federated learning framework with an adaptive trend-awareness module and heterogeneous knowledge alignment, enabling collaborative training of TSFMs across heterogeneous clients. TimeRAF~\cite{TimeRAF} extends TSFM capabilities through retrieval augmentation, dynamically scoring and selecting relevant external time series to improve zero-shot forecasting. TimesFM-ICF~\cite{TimesFM-ICF} draws inspiration from LLMs by enabling in-context adaptation, where prompts enriched with related time series examples allow the model to recover the benefits of domain-specific fine-tuning at inference time.
TTM~\cite{TTM} introduces specialized architectural enhancements to handle heterogeneous datasets with small models, including adaptive patching, diverse resolution sampling, and resolution prefix tuning.
TimeDiT~\cite{TimeDiT} replaces autoregressive generation with a diffusion-based paradigm, addressing the limitations of conventional TSFMs in modeling complex distributions, supporting diverse predictions, integrating external knowledge, and improving efficiency for long-sequence generation. ROSE~\cite{ROSE} proposes the Time Series Register (TS-Register), a memory-like module that stores domain-specific information during pre-training and adaptively injects relevant register tokens for cross-domain transfer.

\subsection{Task Configuration}
Time series analysis encompasses diverse tasks, including forecasting, imputation, anomaly detection, and classification. 
While most existing TSFMs concentrate on a single task—primarily forecasting due to its wide applicability and abundant datasets—the real challenge lies in unifying multiple tasks within a shared framework. Such multi-task configurations are critical for capturing common representations, and enabling more versatile applications.

\textbf{Multi-task Configuration} Recent TSFMs extend their capabilities across multiple tasks.
Some multi-task TSFMs, like LPTM~\cite{LPTM} and Moment~\cite{Moment}, learn generalizable representations while keeping most parameters shared, fine-tuning only lightweight task-specific heads. UNITS~\cite{UNITS} designs task tokens to help model distinguish predictive and generative tasks.
Others eliminate the need for additional modules to unify diverse tasks:
Timer~\cite{Timer} reformulates forecasting, imputation, and anomaly detection as a single generative task.
TimeDiT~\cite{TimeDiT} introduces a unified mask mechanism that includes a variety of masks to cater to diverse time series scenarios.
While single-task TSFMs focus on domain-specific optimization, multi-task TSFMs advance toward universal temporal intelligence through shared representations and modular adaptability.

\begin{figure}[ht]
  \centering
  \includegraphics[width=0.49\textwidth]{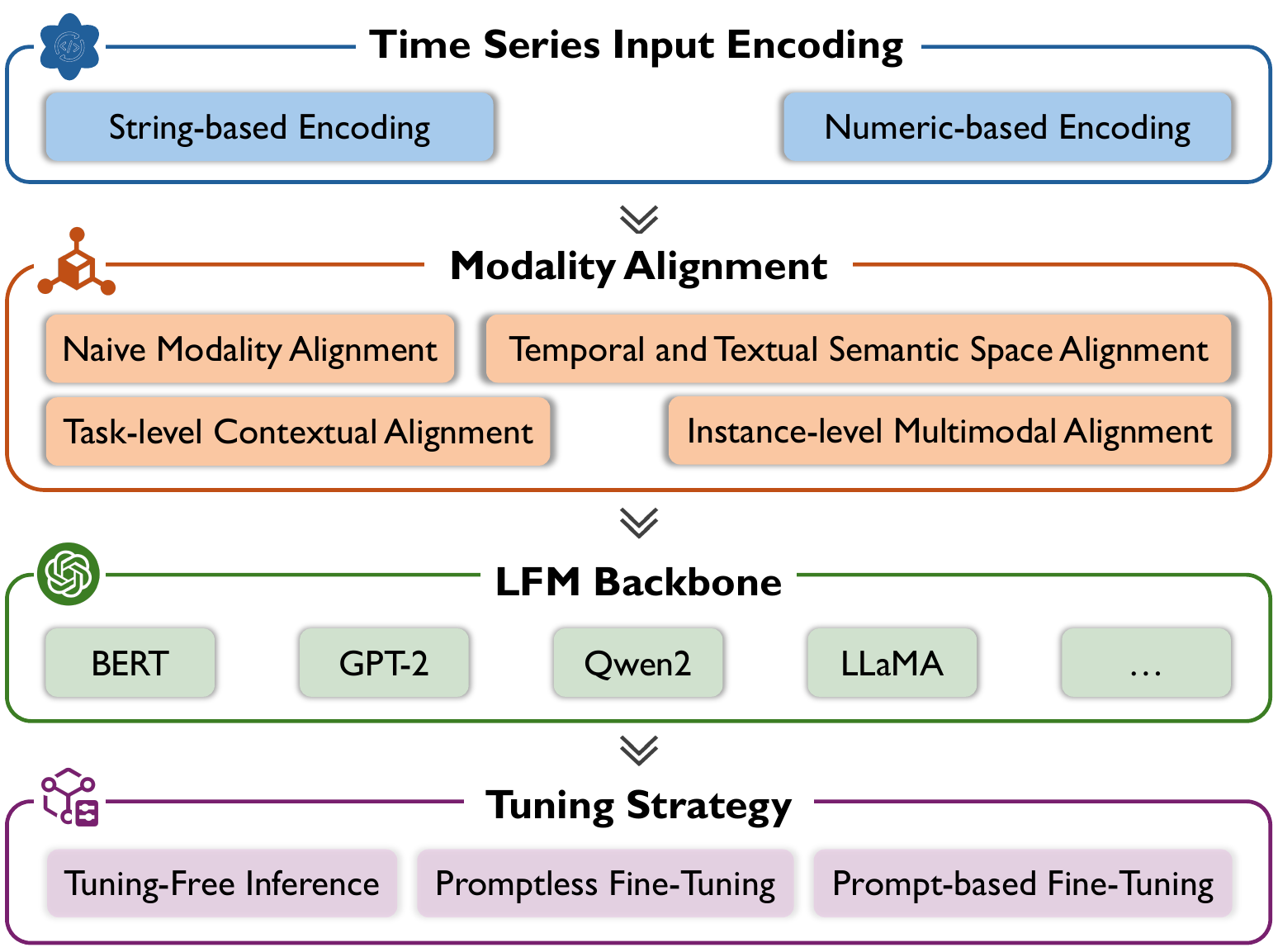}
      \vspace{-2mm}
  \caption{The critical dimensions of adapting language-based foundation models (LFMs) for time series analysis.}
  \label{fig:fig_IV_1}
      \vspace{-4mm}
\end{figure}

\begin{table*}[!ht]
\caption{A summary of Language-Based Foundation Models for Time Series}

\resizebox{\linewidth}{!}{
\begin{threeparttable}

\begin{tabular}{@{}llllllll@{}}
\midrule\midrule

\textbf{Model} &
  \textbf{\begin{tabular}[c]{@{}l@{}}LFM Backbone\\  (Frozen)\end{tabular}} &
  \textbf{Tuning Strategy} &
  \textbf{Tokenization} &
  \textbf{Modality} &
  \textbf{\begin{tabular}[c]{@{}l@{}}Channel \\ Setting\end{tabular}} &
  \textbf{Tasks} &
  \textbf{Code} \\ \midrule
  
LLMTime\cite{LLMTime} &
  GPT-3, LLaMA-2(\ding{55}) &
  Tuning-free Inference &
  String-wise &
  Single &
  Independence &
  F(r) &
  \href{https://github.com/ngruver/llmtime}{Link} \\
PromptCast\cite{PromptCast} &
  BERT, Bart, etc.(\ding{55}) &
  Tuning-free Inference &
  String-wise &
  Multiple (Task) &
  Independence &
  F(o) &
  \href{https://github.com/HaoUNSW/PISA}{Link} \\
SIGLLM\cite{SIGLLM} &
  GPT-2(\ding{55}) &
  Tuning-free Inference &
  String-wise &
  Multiple (Task) &
  Independence &
  D &
  \href{https://github.com/sintel-dev/sigllm}{Link} \\
TableTime\cite{TableTime} &
  Llama-3.1(\ding{55}) &
  Tuning-free Inference &
  String-wise &
  Multiple (Task) &
  Mixing &
  C &
  \href{https://github.com/realwangjiahao/TableTime}{Link} \\
LSTPrompt\cite{LSTPrompt} &
  GPT4(\ding{55}) &
  Tuning-free Inference &
  String-wise &
  Multiple (Task) &
  Independence &
  F(o) &
  \href{https://github.com/AdityaLab/lstprompt}{Link} \\
CALF\cite{CALF} &
  GPT-2(\ding{51}) &
  Promptless Fine-tuning &
  Point-wise &
  Single &
  Independence &
  F(r) &
  \href{https://github.com/Hank0626/CALF}{Link} \\
ISTS-PLM\cite{ISTS-PLM} &
  GPT-2, BERT(\ding{51}) &
  Promptless Fine-tuning &
  Point-wise &
  Single &
  Mixing &
  C &
  \href{https://github.com/usail-hkust/ISTS-PLM}{Link} \\
LLM4TS\cite{LLM4TS} &
  GPT-2(\ding{51}) &
  Promptless Fine-tuning &
  Patch-wise &
  Single &
  Independence &
  F(o) &
  N/A \\
FPT\cite{FPT} &
  GPT-2(\ding{51}) &
  Promptless Fine-tuning &
  Patch-wise &
  Single &
  Independence &
  F(o)/C/D/I &
  \href{https://github.com/DAMO-DI-ML/One_Fits_All}{Link} \\
aLLM4TS\cite{aLLM4TS} &
  GPT-2(\ding{51}) &
  Promptless Fine-tuning &
  Patch-wise &
  Single &
  Independence &
  F(o)/D &
  \href{https://github.com/yxbian23/aLLM4TS}{Link} \\
AnomalyLLM\cite{AnomalyLLM} &
  GPT-2(\ding{51}) &
  Promptless Fine-tuning &
  Patch-wise &
  Single &
  Independence &
  D &
  \href{https://github.com/fly-orange/AnomalyLLM}{Link} \\
MedualTime\cite{MedualTime} &
  GPT-2(\ding{51}) &
  Promptless Fine-tuning &
  Patch-wise &
  Multiple (Instance) &
  Mixing &
  C &
  \href{https://github.com/start2020/MedualTime}{Link} \\
LeRet\cite{LeRet} &
  LLaMA(\ding{55}) &
  Promptless Fine-tuning &
  Patch-wise &
  Multiple (Instance) &
  Independence &
  F(o) &
  \href{https://github.com/hqh0728/LeRet}{Link} \\
TimeCAP\cite{TimeCAP} &
  GPT-4, BERT(\ding{55}) &
  Prompt-based Fine-tuning &
  String-wise &
  Multiple (Instance) &
  Independence &
  E &
  \href{https://github.com/geon0325/TimeCAP}{Link} \\
ChatTime\cite{ChatTime} &
  LLaMA-2(\ding{51}) &
  Prompt-based Fine-tuning &
  String-wise &
  Multiple (Task) &
  Independence &
  F(o)/QA &
  \href{https://github.com/ForestsKing/ChatTime}{Link} \\
From-News\cite{From-News} &
  GPT-4 Turbo(\ding{51}) &
  Prompt-based Fine-tuning &
  String-wise &
  Multiple (Instance) &
  Independence &
  F(r) &
  \href{https://github.com/ameliawong1996/From_News_to_Forecast}{Link} \\
TEST\cite{TEST} &
  GPT-2, BERT, etc.(\ding{55}) &
  Prompt-based Fine-tuning &
  Patch-wise &
  Multiple (Task) &
  Mixing &
  F(o)/C &
  N/A \\
Time-LLM\cite{Time-LLM} &
  LLaMA(\ding{55}) &
  Prompt-based Fine-tuning &
  Patch-wise &
  Multiple (Task) &
  Independence &
  F(o) &
  \href{https://github.com/KimMeen/Time-LLM}{Link} \\
TEMPO\cite{TEMPO} &
  GPT-2(\ding{51}) &
  Prompt-based Fine-tuning &
  Patch-wise &
  Single &
  Independence &
  F(o) &
  \href{https://github.com/dc-research/tempo}{Link} \\
S2IP-LLM\cite{S2IP-LLM} &
  GPT-2(\ding{51}) &
  Prompt-based Fine-tuning &
  Patch-wise &
  Multiple (Task) &
  Independence &
  F(o) &
  \href{https://github.com/panzijie825/S2IP-LLM}{Link} \\
NuwaTS\cite{NuwaTS} &
  GPT-2(\ding{51}) &
  Prompt-based Fine-tuning &
  Patch-wise &
  Multiple (Task) &
  Independence &
  I &
  \href{https://github.com/Chengyui/NuwaTS}{Link} \\
Time-FFM\cite{Time-FFM} &
  GPT-2(\ding{55}) &
  Prompt-based Fine-tuning &
  Patch-wise &
  Multiple (Task) &
  Independence &
  F(o) &
  \href{https://github.com/CityMind-Lab/NeurIPS24-Time-FFM}{Link} \\
UniTime\cite{UniTime} &
  GPT-2(\ding{51}) &
  Prompt-based Fine-tuning &
  Patch-wise &
  Multiple (Task) &
  Independence &
  F(o) &
  \href{https://github.com/liuxu77/UniTime}{Link} \\
AutoTimes\cite{AutoTimes} &
  LLaMA(\ding{55}) &
  Prompt-based Fine-tuning &
  Patch-wise &
  Multiple (Task) &
  Independence &
  F(o) &
  \href{https://github.com/thuml/AutoTimes}{Link} \\
HiTime\cite{HiTime} &
  LLaMA-3.1(\ding{51}) &
  Prompt-based Fine-tuning &
  Patch-wise &
  Multiple (Instance) &
  Mixing &
  C &
  \href{https://github.com/Xiaoyu-Tao/HiTime}{Link} \\
GPT4MTS\cite{GPT4MTS} &
  GPT-2(\ding{51}) &
  Prompt-based Fine-tuning &
  Patch-wise &
  Multiple (Instance) &
  Independence &
  F(o) &
  \href{https://github.com/Flora-jia-jfr/GPT4MTS-Prompt-based-Large-Language-Model-for-Multimodal-Time-series-Forecasting}{Link} \\
ExoLLM\cite{ExoLLM} &
  NA(\ding{51}) &
  Prompt-based Fine-tuning &
  Patch-wise &
  Multiple (Task) &
  Independence &
  F(o) &
  \href{https://github.com/h505023992/ExoLLM}{Link} \\
InstructTime\cite{InstructTime} &
  GPT-2(\ding{51}) &
  Prompt-based Fine-tuning &
  Patch-wise &
  Multiple (Task) &
  Independence &
  C &
  \href{https://github.com/Mingyue-Cheng/InstructTime}{Link} \\
FSCA\cite{FSCA} &
  GPT-2(\ding{51}) &
  Prompt-based Fine-tuning &
  Patch-wise &
  Multiple (Task) &
  Independence &
  F(o)/C &
  \href{https://github.com/tokaka22/ICLR25-FSCA}{Link} \\
STEM-LTS\cite{STEM-LTS} &
  GPT-2(\ding{51}) &
  Prompt-based Fine-tuning &
  Patch-wise &
  Multiple (Task) &
  Independence &
  F(o) &
  \href{https://github.com/DataLabatom/STEM-LTS}{Link} \\
LangTime\cite{LangTime} &
  Qwen2(\ding{51}) &
  Prompt-based Fine-tuning &
  Patch-wise &
  Multiple (Task) &
  Independence &
  F(o) &
  \href{https://github.com/niuwz/LangTime}{Link}

\\ \bottomrule
\end{tabular}

\begin{tablenotes}
\footnotesize
\item Note:
\textbf{\ding{55}} refers to freeze LFM backbone while \textbf{\ding{51}} refers to fine-tune LFM backbone.
\textbf{Single} denotes only time series input; \textbf{Multiple (Task)} denotes time series with global contextual text; 
\textbf{Multiple (Instance)} denotes time series with sample-specific text.
Task setting: \textbf{F(o)} is Point  Forecast; \textbf{F(r)} is Probabilistic Forecast; \textbf{I} is Imputation; \textbf{D} is Anomaly Detection; \textbf{C} is Classification; \textbf{E} is Event Prediction. \textbf{N/A} denotes \textit{Not Applicable}.
\end{tablenotes}
\end{threeparttable}
}

\label{tab:tab_IV_1}
\end{table*}

\section{Language-based Foundation Model For Time Series}
\label{sec:text}

Unlike TSFMs, language-based foundation models are pre-trained on massive textual corpus, which poses unique challenges on adapting them to time series tasks. 
This section summarizes three key challenges in this adaptation:
first, time series input encoding, which inquires input compatibility by transforming time series data into representations that LFMs can effectively perceive and process; 
second, modality alignment, which inquires knowledge transfer from language model to time series tasks;
third, tuning strategies, which inquires a balance between model performance and computational efficiency.
We review and categorize existing solutions, highlighting those efforts that unlock the potential of LFMs in time series analysis.
Figure~\ref{fig:fig_IV_1} summarizes the content of this section.

\subsection{Time Series Input Encoding}
Time series input encoding can be broadly categorized into two types. \textit{String-based encoding} converts numerical sequences into discrete text tokens, enabling direct use of LFMs’ pre-trained linguistic capabilities, but it often introduces quantization errors and struggles to preserve fine-grained temporal patterns. In contrast, \textit{numeric-based encoding} operates directly on numeric values to better capture temporal dynamics, yet it requires to align time series with language space.

\textbf{String-based Encoding} This approach first converts numerical time series values into string representations, enabling the reuse of language model tokenizers and vocabularies and  allowing seamless integration between string-formatted time series and textual prompts. However, when an LFM’s tokenizer splits consecutive digits into separate tokens, it may disrupt temporal relationships. To address this, LLMTime~\cite{LLMTime} introduces spaces between digits, prompting GPT-3 to treat each digit as an individual token, improving performance. Additionally, since LFM tokenizers are not optimized for numerical data, capturing dynamic temporal patterns remains challenging. Some methods incorporate prompts to provide context and enhance LFM's reasoning capacity in time series analysis~\cite{PromptCast}.~\cite{spathis2024first} argues that numerical string encoding does not fully harness the potential of time series data. The true value of time series data can only be realized when mapped to a more meaningful representation suited to its inherent nature, rather than being forced into a text-based format.

\textbf{Numeric-based Encoding} Instead of converting values into strings, this approach directly tokenizes raw numerical sequences, often employing patch-wise or point-wise tokenization to generate time series representation.
For patch-wise tokenization, the encoding of univariate time series usually begins with reversible instance normalization~\cite{RevIN} to standardize the data and mitigate distribution shifts. Then, patching~\cite{PatchTST} aggregates adjacent time points into patches, preserving local semantics. Finally, a patch embedder, typically a linear layer, transforms these patches into embeddings compatible with the LFMs’ input space. For multivariate time series, many methods decompose the data into univariate sequences for independent patching. Other methods like like HiTime~\cite{HiTime}, MedualTime~\cite{MedualTime} do not treat variables independently but preserve multivariate input to capture cross-variable relationships during encoding. 

To better adapt LFMs to time series data, recent methods enhance input representations with additional structural, statistical, and semantic information. Some methods incorporate positional and timestamp embeddings to help LFMs capture temporal structures. AutoTimes~\cite{AutoTimes} aligns segment tokens by converting timestamps into position embeddings.
NuwaTS~\cite{NuwaTS} uses statistical and missing embeddings to encode data properties for imputation.
UniTime~\cite{UniTime} applies masking on time series tokens to improve prediction in cross-domain training.
TEMPO~\cite{TEMPO}, S2IP-LLM~\cite{S2IP-LLM} and STEM-LTS~\cite{STEM-LTS} decompose time series into seasonal and trend components before patching, capturing patterns that LFM’s attention mechanism struggles to identify.
TimeCAP~\cite{TimeCAP} concatenates its time-series patch embedding of each channel with the text embedding and applies multi-head self-attention to capture interactions between temporal segments and textual tokens.

\begin{figure}[ht]
\vspace{-3mm}
  \centering
  \includegraphics[width=0.49\textwidth]{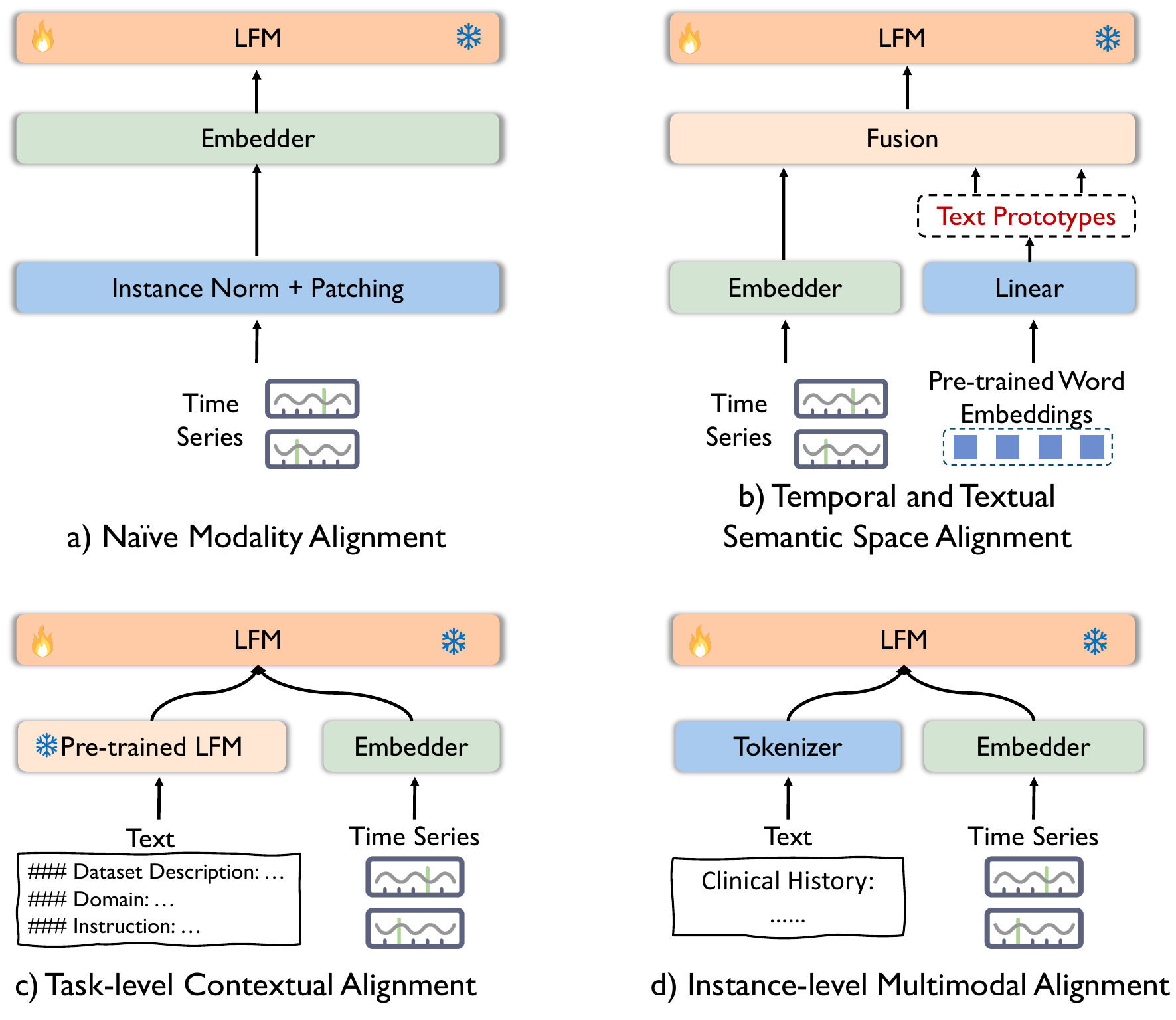}
      \vspace{-2mm}
  \caption{Illustration of four modality alignment strategies that map time series to the semantic space of language-based foundation models (LFMs).}
  \label{fig:fig_IV_2}
      \vspace{-4mm}
\end{figure}
\subsection{Modality Alignment}

The adaptation of LFMs to time series analysis is fundamentally challenged by the modality gap: textual data are discrete, symbolic, and organized by linguistic syntax, whereas time series are continuous numeric signals shaped by temporal dependencies and convey patterns rather than explicit semantics. This structural and semantic discrepancy makes it difficult for LFMs to perceive temporal dynamics or extract task-relevant semantics from raw time series.  Modality alignment aims to align time series with the semantic space of textual knowledge so that LFMs’ pre-trained capabilities can be effectively transferred to time series tasks. To achieve this, several strategies have been explored (as shown in Figure~\ref{fig:fig_IV_2}): \textit{Naive Modality Alignment} employs straightforward transformations to bridge the gap; \textit{Temporal and Textual Semantic Space Alignment} aligns temporal representation and textual tokens in language space; \textit{Task-level Contextual Alignment} provides shared contextual guidance that steers LFMs toward reasoning on time series task; while \textit{Instance-level Multimodal Alignment} integrates sample-specific paired modalities to exploit complementary information and achieve cross-modal fusion.


\textbf{Naive Modality Alignment}
Some single-modal approaches with only time series input achieve modality alignment by simply adjusting the dimension of the time series embedding to match the LFM input space~\cite{FPT,LLM4TS} or directly converting time series into strings and feeding them into the LFM without any prompts~\cite{LLMTime}. While these methods are straightforward and have been empirically validated, they overlook the distributional discrepancy between time series and language tokens, which can lead to suboptimal model performance.

\textbf{Temporal and Textual Semantic Space Alignment} 
Although both numerical time series and natural language are sequential, their embedding spaces differ significantly due to distributional discrepancies. This modality gap makes it challenging for LFMs, pre-trained on textual tokens, to interpret time series embeddings effectively. To address this, TEST~\cite{TEST} employs feature-wise contrastive learning to map time series embeddings onto pivot text prototype vectors within the LFM’s semantic space. 
Time-LLM~\cite{Time-LLM} reprograms the patch embeddings with text prototypes from pre-trained word embeddings by multi-head cross-attention to achieve modality alignment.
Similarly, CALF~\cite{CALF} aligns the input distribution of time series with LFM word embeddings via a cross-attention module.
By explicitly mapping temporal embeddings to the LFM’s semantic space, these alignment approaches bridge the modality gap, allowing LFMs to leverage their pre-trained linguistic knowledge for enhanced comprehension and reasoning over time series data.

\textbf{Task-level Contextual Alignment}
This type of alignment leverages global contextual information—such as task instructions, domain knowledge, and prior expertise—to guide LFMs in understanding both the time series task and data. In practice, such information is predominantly injected in the form of textual prompts, serving as contextual guidance for effective reasoning on each downstream task.
Textual prompts, such as template engineering and chain-of-thought, are intuitive and easy to design. Some studies directly integrate prompts with textualized time series  to activate LFM reasoning~\cite{LSTPrompt}, while others prepend tokenized textual prompts to continuous time series embeddings~\cite{AutoTimes}, aligning them in the LFM’s latent space with minimal computational complexity.
The quality of prompts is crucial for guiding LFMs and a well-structured prompt can provide LFMs with essential context and direction to ensure that the model output aligns with expectations. Typical prompts include:
(1) Task instruction – defines clear instructions for processing time series~\cite{Time-LLM}.
(2) Domain description – provides background to help identify domain-specific patterns~\cite{UniTime}.
(3) Dataset description – conveys essential metadata or contextual attributes of the dataset~\cite{LangTime}.
(4) Prior knowledge – enhances inference with domain expertise~\cite{HiTime}.

\textbf{Instance-level Multimodal Alignment}
This type of alignment focuses on sample-specific paired modalities, such as ECG signals with corresponding clinical notes~\cite{METS} or financial time series with related news articles. Unlike task-level alignment, which mainly relies on global prompts, instance-level alignment integrates complementary modalities directly at the sample level. These approaches span from direct attention modeling to lightweight prompt injection, as well as more advanced agent-driven or adapter-based interactions.


HiTime~\cite{HiTime} aligns time series and label-included text descriptions through a cross-attention mechanism, enabling both coarse-grained and fine-grained fusion between temporal and textual embeddings.
GPT4MTS~\cite{GPT4MTS} creates textual summaries for numerical media coverage. They extracts textual embeddings with BERT, which are prepended to temporal embeddings to form multimodal inputs.
Similarly, TimeCAP~\cite{TimeCAP} utilizes a first agent to generate sample-specific textual summaries from time series and a second agent to leverage these summaries together with the raw series to make event predictions.
LeRet~\cite{LeRet} constructs sample-specific textual descriptors (e.g., trend, period, stability) for each time series, yielding language-empowered representations for forecasting.
\cite{From-News} integrates time series–news pairs through an LFM-agent framework, where agents dynamically filter relevant news and pair it with time series data, enabling context-aware multimodal forecasting.
MedualTime~\cite{MedualTime} designs a dual-adapter language model for medical signal-clinical report pair where either modality can serve as the primary modality while being enhanced by the other. 

\begin{figure}[ht]
  \centering
  \includegraphics[width=0.49\textwidth]{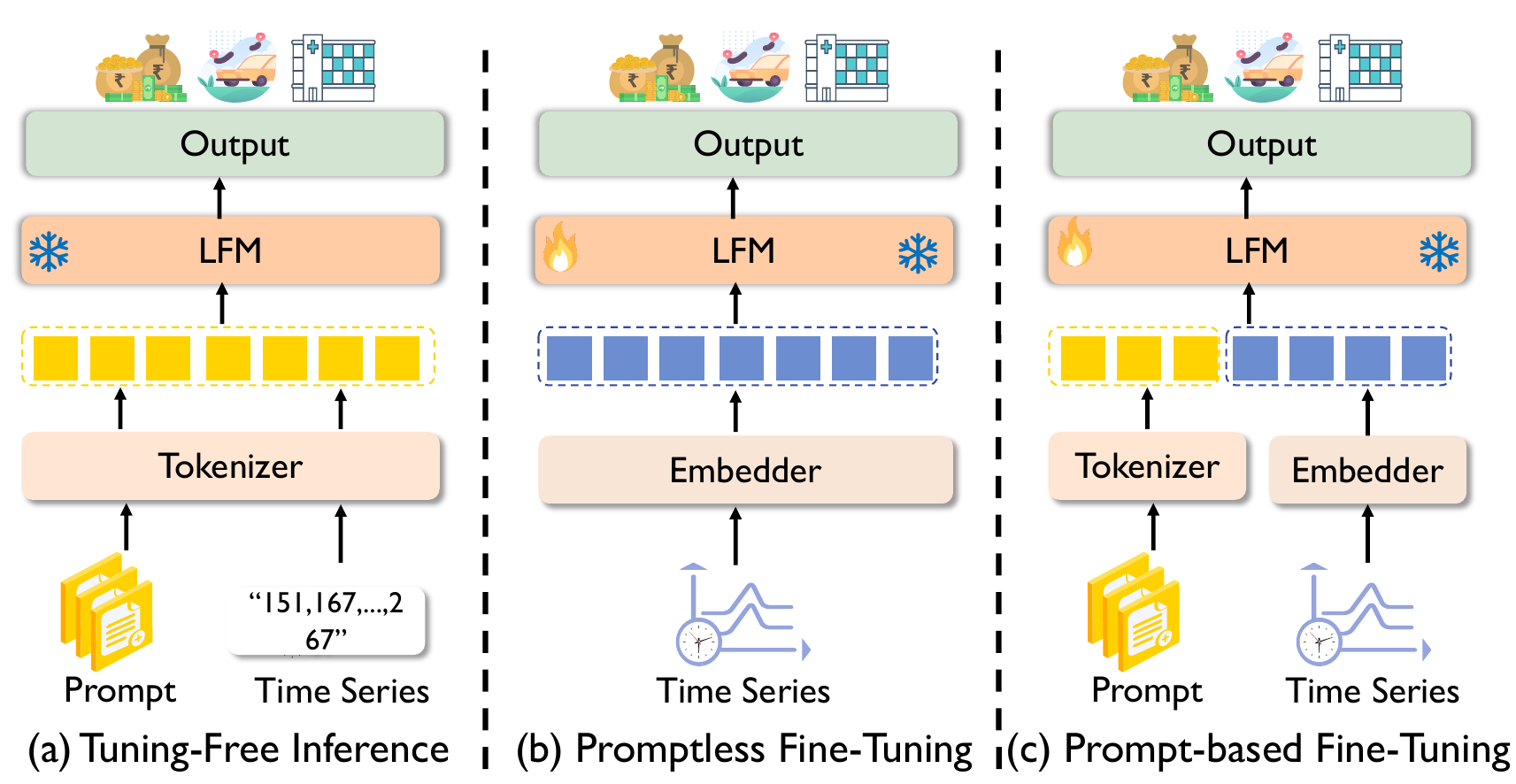}
      \vspace{-2mm}
  \caption{Illustrations of tuning strategies in adapting LFMs for time series.}
  \label{fig:fig_IV_3}
      \vspace{-4mm}
\end{figure}
\subsection{Tuning Strategy}

After pre-training, LFMs acquire general-purpose capabilities applicable to a wide range of tasks. To adapt them for time series analysis, existing methods explore different tuning strategies that balance task-specific optimization, efficiency, and flexibility. These strategies can be broadly categorized into three types (as shown in Figure~\ref{fig:fig_IV_3}):  
\textit{Tuning-Free Inference}, which directly applies pre-trained LFMs to time series tasks without updating any parameters.  
\textit{Promptless Fine-Tuning}, which follows the conventional "pre-train and fine-tune" paradigm by directly fine-tuning LFMs on time series data without relying on prompts.  
\textit{Prompt-based Fine-Tuning}, which constructs hard or soft prompts to guide the fine-tuning  for better task adaptation.

\textbf{Tuning-Free Inference} Its typical pipeline involves converting time series into textual input, designing structured prompts and concatenating the input and prompt to activate the LFM’s reasoning capabilities.
Prompt-Cast~\cite{PromptCast} pioneers the transformation of time series input and output into a question-answering prompt, enabling LFMs to perform forecasting.
TableTime~\cite{TableTime} converts multivariate time series into tabular format, serializes and integrates them into a prompt with contextual information, neighboring knowledge, and task decomposition for zero-shot classification.
LLMTime~\cite{LLMTime} encodes time series as strings of separate numerical digits and feeds them into the LFM without a prompt, achieving good zero-shot forecasting.
SIGLLM~\cite{SIGLLM} follows a similar "time series-to-text" approach as LLMTime but incorporates prompts to assess the LFM’s ability to detect time series anomalies.
Overall, these approaches demonstrate that converting time series into textual representations allows LFMs to leverage their reasoning and language modeling capabilities for time series tasks. Different prompt designs and input formats highlight complementary advantages in enabling zero-shot learning.

\textbf{Promptless Fine-tuning} In contrast to tuning-free inference, which leverages prompts without parameter updates, this strategy discards prompts and usually modifies the input/output layers, updating LFMs for fine-tuning. While full fine-tuning achieves strong performance on large datasets, it struggles with limited data and risks catastrophic forgetting. To mitigate this, most studies adopt parameter-efficient fine-tuning (PEFT), updating only selected LFM components.
FPT~\cite{FPT} and AnomalyLLM~\cite{AnomalyLLM} fine-tune embedding and normalization layers while freezing self-attention and feedforward layers.
CALF~\cite{CALF} applies Low-Rank Adaptation~\cite{LoRA} to update low-rank matrices in self-attention blocks.
MedualTime~\cite{MedualTime} employs adapter tuning for efficient multimodal fusion.
Beyond these one-stage fine-tuning approaches, some methods adopt a two-stage fine-tuning process. LLM4TS~\cite{LLM4TS} first applies PEFT to align LFM with time series data, followed by full fine-tuning for task-specific adaptation.
aLLM4TS~\cite{aLLM4TS} takes the reverse steps, starting with full fine-tuning, then refining LFM with PEFT to balance generalization and efficiency.

\textbf{Prompt-based Fine-tuning} 
This strategy exploits prompt design to fine-tune LFMs for time-series tasks. Hard prompts provide explicit, human-aligned instructions: Time-LLM~\cite{Time-LLM} and HiTime~\cite{HiTime} integrate dataset context, task definitions, and domain knowledge; UniTime~\cite{UniTime} uses domain instructions to mitigate domain confusion; AutoTimes~\cite{AutoTimes} leverages lookback–forecast windows as in-context demonstrations. However, TEST~\cite{TEST} argues that hard prompts diverge from time-series embeddings and instead introduces trainable soft prompts. Other works extend this direction: GPT4MTS~\cite{GPT4MTS} treats coupled text embeddings as soft prompts, Time-FFM~\cite{Time-FFM} aligns time-series patches with corpus tokens through adaptive prompt design, TEMPO~\cite{TEMPO} builds a soft prompt pool encoding historical temporal effects, S2IP-LLM~\cite{S2IP-LLM} retrieves semantic anchors from the pre-trained space, and NuwaTS~\cite{NuwaTS} applies prefix-tuning with domain-specific embeddings. Broadly, these methods fall into hard prompts (intuitive, human-readable) and soft/adaptive prompts (flexible, pattern-oriented). Regarding the LFM backbone, some works freeze it and fine-tune lightweight modules for efficiency, while others directly fine-tune the backbone to strengthen time-series reasoning.

While tuning-free inference is highly efficient and requires no additional training, it relies solely on LFMs' pre-trained knowledge and well-structured, manually crafted prompts to guide LFMs' outputs effectively. Despite leveraging the generative strengths of LFMs, this approach does not update the LFMs' internal knowledge to better understand time series data. Prompt-less fine-tuning generally achieves higher task-specific accuracy by updating a small subset of parameters. However, without explicit prompts, LFMs may struggle to grasp task context, potentially impairing their reasoning capabilities. Prompt-based fine-tuning is widely used, combining the benefits of prompting and fine-tuning for better adaptability.
\begin{figure}[ht]
  \centering
  \includegraphics[width=0.49\textwidth]{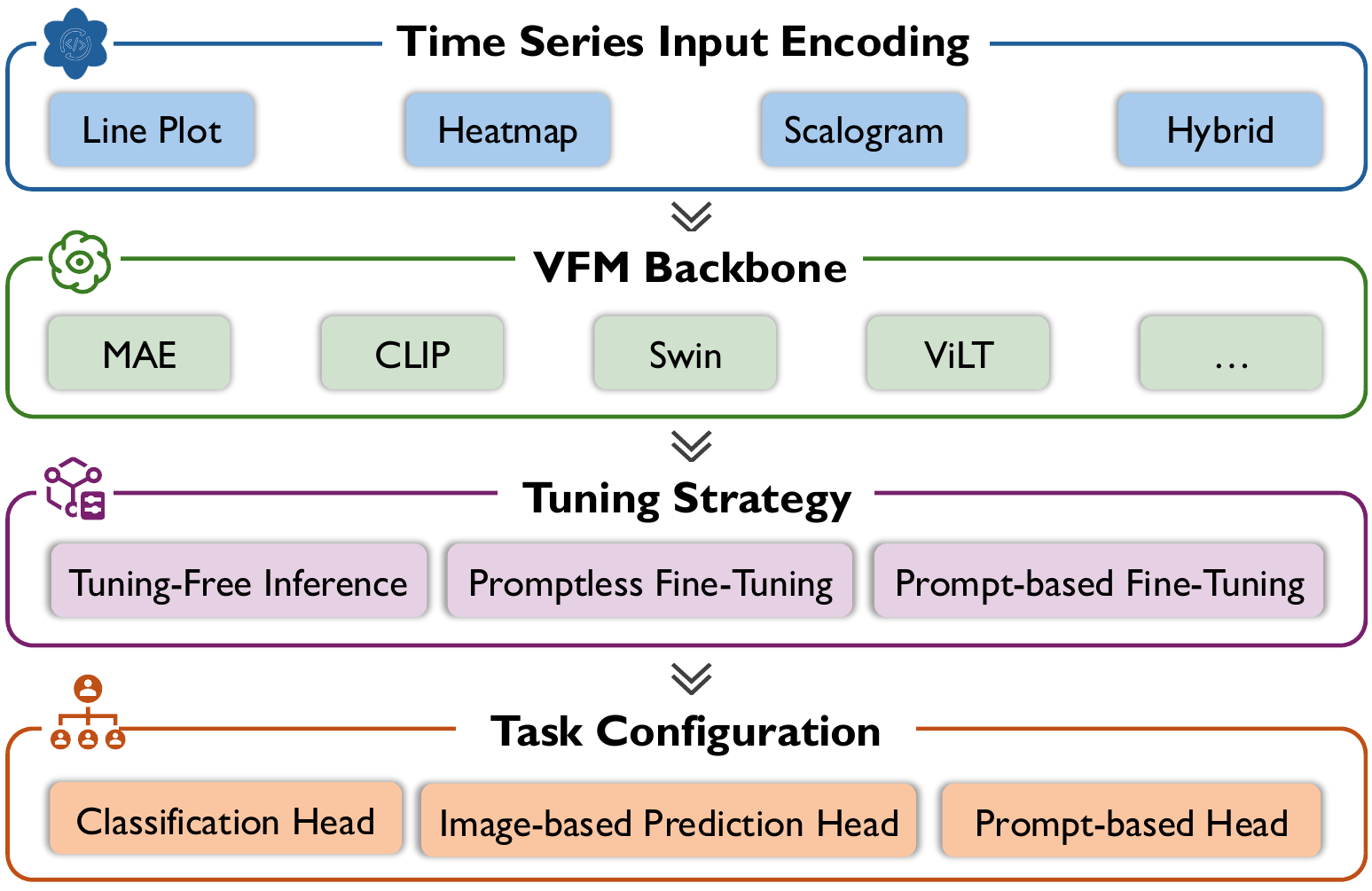}
  \caption{The critical dimensions of adapting vision-based foundation models (VFMs) for time series analysis.}
  \label{fig:fig_V_1}
      \vspace{-4mm}
\end{figure}
\begin{table*}[!ht]
\caption{A summary of Vision-Based Foundation Models For Time Series}
\resizebox{\linewidth}{!}{
\begin{threeparttable}

\begin{tabular}{@{}llllllll@{}}
\midrule \midrule

\textbf{Model} &
  \textbf{Backbone} &
  \textbf{Tuing Strategy} &
  \textbf{Imaging} &
  \textbf{Modality} &
  \textbf{\begin{tabular}[c]{@{}l@{}}Channel\\  Setting\end{tabular}} &
  \textbf{Tasks} &
  \textbf{Code} \\ \midrule
  
ITF-TAD\cite{ITF-TAD} &
  PatchCore &
  Tuning-free Inference &
  Scalogram &
  Single &
  Independence &
  D &
  {\color[HTML]{242424} N/A} \\
VLM4TS\cite{VLM4TS} &
  CLIP &
  Tuning-free Inference &
  LinePlot &
  Multiple (Task) &
  Independence &
  D &
  \href{https://github.com/ZLHe0/VLM4TS}{Link} \\
VisionTS\cite{VisionTS} &
  MAE &
  Tuning-free Inference &
  Heatmap &
  Single &
  Independence &
  F(o) &
  \href{https://github.com/ Keytoyze/VisionTS}{Link} \\
ViTST\cite{ViTST} &
  Swin &
  Promptless Fine-tuning &
  LinePlot &
  Single &
  Mixing &
  C &
  \href{https://github.com/Leezekun/ViTST}{Link} \\
ViTime\cite{ViTime} &
  ViT &
  Promptless Fine-tuning &
  Heatmap &
  Single &
  Mixing &
  F(o) &
  \href{https://github.com/IkeYang/ViTime}{Link} \\
DMMV\cite{DMMV} &
  MAE &
  Promptless Fine-tuning &
  Heatmap &
  Multiple (Instance) &
  Independence &
  F(o) &
  {\color[HTML]{242424} N/A} \\

  VisionTS++\cite{VisionTS++} &
  MAE &
  Promptless Fine-tuning &
  Heatmap &
  Single &
  Mixing &
  F(r) &
  \href{https://github.com/HALF111/VisionTSpp}{Link} \\
LDM4TS\cite{LDM4TS} &
  U-Net &
  Promptless Fine-tuning &
  Hybrid &
  Multiple (Instance) &
  Mixing &
  F(r) &
  {\color[HTML]{242424} N/A} \\
OccamVTS\cite{OccamVTS} &
  MAE, CLIP &
  Promptless Fine-tuning &
  Hybrid &
  Multiple (Instance) &
  Mixing &
  F(o) &
  {\color[HTML]{242424} N/A} 

   \\
VLM-TSC\cite{VLM-TSC} &
  LLaVA &
  Prompt-based Fine-tuning &
  LinePlot &
  Multiple (Instance) &
  Mixing &
  C &
  \href{https://github.com/vinayp17/VLM_TSC}{Link} \\
Time-VLM\cite{Time-VLM} &
  ViLT &
  Prompt-based Fine-tuning &
  Hybrid &
  Multiple (Instance) &
  Mixing &
  F(o) &
  \href{https://github.com/ CityMind-Lab/ICML25-TimeVLM}{Link}
  
  \\ \midrule \midrule
  
\end{tabular}

\begin{tablenotes}
\footnotesize
\item Note: 
\textbf{Single} denotes only image based time series input; \textbf{Multiple (Task)} denotes image based time series with global contextual textual information; 
\textbf{Multiple (Instance)} denotes image based time series with paired numerical time series or text. Task setting: \textbf{F(o)} is Point  Forecast; \textbf{F(r)} is Probabilistic Forecast; \textbf{D} is Anomaly Detection; \textbf{C} is Classification.
\end{tablenotes}
\end{threeparttable}

}
\label{tab:tab_V_1}
\end{table*}
\section{Vision-based Foundation Models For Time Series}
\label{sec:image}
Vision-based Foundation Models (VFMs) are pre-trained on large-scale visual datasets or multimodal corpora containing paired images and text. Compared to LFMs, VFMs presents both shared and unique adaptation challenges for time series tasks. In this section, we outline four critical aspects of adapting VFMs for time series analysis (as shown in Figure \ref{fig:fig_V_1}): (i) image-based encoding, which converts temporal data into image-like representations compatible with visual architectures; (ii) backbone adaptation, which exploits VFMs’ spatial pattern recognition to capture temporal dynamics; (iii) tuning strategies, which adapt visual backbones to time series tasks while balancing performance and efficiency; and (iv) task recovery, which maps visual outputs back to temporal objectives. We categorize recent studies by their solutions in each aspect and analyze their relative advantages and drawbacks, offering deeper insights into ongoing technical progress.

\begin{figure}[ht]
  \centering
  \includegraphics[width=0.49\textwidth]{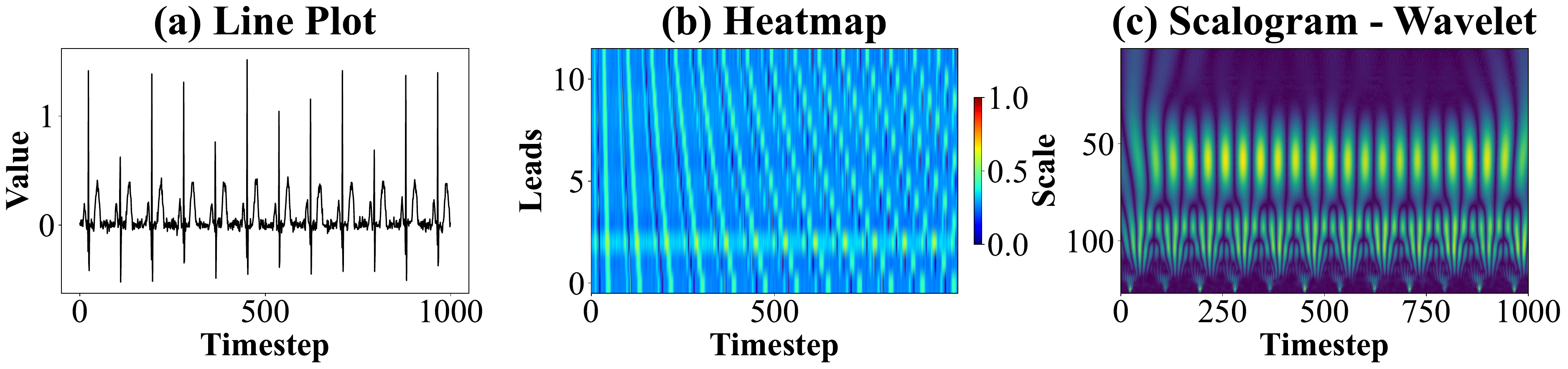}
      \vspace{-2mm}
  \caption{Illustration of representative strategies for converting time series into image-based representations.}
  \label{fig:fig_V_2}
      \vspace{-4mm}
\end{figure}
\subsection{Time Series Input Encoding}
Transforming time series into images provides a natural bridge for leveraging pre-trained vision backbones. We categorize existing approaches into four categories of image representations—line plots, heatmaps, spectrograms, and hybrid encodings (as shown in Figure \ref{fig:fig_V_2})—and discuss their advantages, representative methods, and limitations.

\textbf{Line Plot-based Encoding}
Line plots are the most intuitive visualization of time series, highlighting trends, peaks, and anomalies in a format easily interpreted by both humans and models. ViTST~\cite{ViTST} converts each variate into a line sub-plot, arranges them into a grid to involve both local and global contexts. VLM-TSC~\cite{VLM-TSC} also uses clean line plots (without axes or legends) and demonstrates that line plots outperform scatter plots for classification. VLM4TS~\cite{VLM4TS} renders sliding windows into square line plots, aligning one time step per pixel column and fixing y-axis ranges for consistency across windows. While line plots are easy to generate and semantically rich, they are sensitive to plotting conventions, resolution, and scaling, which can introduce inconsistencies across datasets.

\textbf{Heatmap-based Encoding}
Heatmap-style encodings map time series into dense matrices, which naturally align with transformer-style patch embeddings and sometimes allow invertibility back to numeric sequences. VisionTS~\cite{VisionTS} converts time series into images by segmenting them periodically, stacking them into a normalized 2D matrix, and replicating grayscale values into three channels. ViTime~\cite{ViTime} maps sequences into a binary image space with explicit forward and inverse functions, ensuring exact recoverability of time series values from predicted images. DMMV~\cite{DMMV} also adopts MAE-based imaging and introduces BCMASK to reconstruct left/right halves, mirroring forecasting tasks. Heatmap representations thus exploit patch-level processing in transformers and support recovery for forecasting, but they may oversimplify cross-variable correlations and can be sensitive to segmentation.

\textbf{Scalogram-based Encoding}
Scalograms emphasize the joint representation of temporal and frequency content, which is especially useful for anomaly detection. ITF-TAD~\cite{ITF-TAD} converts multivariate signals into scalograms via continuous wavelet transforms (CWT), aggregates them, and packs them into RGB images. This design explicitly preserves spectral characteristics and allows model to detect deviations in frequency patterns that correspond to anomalies. However, spectrogram-based encodings are generally non-invertible, making them less suited for forecasting or imputation tasks that require recovering numeric sequences.

\textbf{Hybrid Encoding}
Hybrid encodings aim to combine multiple signals—temporal, frequency, or semantic—to improve generalization. 
Time-VLM~\cite{Time-VLM} encodes time series via three modalities: three-channel images from multi-scale convolutions with frequency and periodic encodings, numerical patches for retrieval, and textual descriptions (e.g., statistics and domain context) for semantic extraction. By fusing these modalities, the model exploits complementary information sources and improves robustness under domain shift. Such hybrid approaches are flexible and powerful, but they introduce higher design complexity and rely on careful multimodal alignment to fully realize their advantages.

In summary, line plots provide interpretability but are highly sensitive to visual conventions; heatmaps align well with transformer backbones and can enable recoverability but depend on segmentation design; scalograms preserve frequency information but lack invertibility; and hybrid encodings maximize flexibility by fusing multiple modalities but at the cost of greater complexity. These trade-offs underscore that image encoding fundamentally shapes what temporal information is preserved, how it aligns with the backbone, and what downstream tasks are feasible.

\subsection{Backbone Adaptation}
Image based representations allows time series to be processed by vision models. In this section, we focus on backbone-level adaptations, analyzing how existing works repurpose vision-based foundation models (VFMs) for imaged time series tasks, why these backbones are chosen, and how temporal characteristics are implicitly or explicitly considered.

\textbf{CNN-based Backbone}
Early attempts have employed CNNs for their strong ability to capture local spatial patterns. A representative example is ITF-TAD~\cite{ITF-TAD}, which uses a frozen ResNet to extract features from scalogram images. By relying on the convolutional bias toward frequency–time textures, CNN-based VFMs offer robustness and efficiency for anomaly detection. However, their limited capacity for long-range dependency modeling constrains their broader use in forecasting tasks.

\textbf{Vision Transformer-based Backbone}
Transformer-based models have been more widely adopted as VFMs backbones because their patch-based representations naturally align with temporal segmentation. To compensate for the absence of explicit temporal inductive bias, existing approaches augment them with techniques like decoders~\cite{ViTime} or multi-view strategies~\cite{DMMV}. ViTST~\cite{ViTST} fine-tunes Swin on line-plot images of irregularly sampled time series, exploiting self-attention to capture cross-variable dependencies. VisionTS~\cite{VisionTS} reformulates forecasting as masked image reconstruction with MAE, reinterpreting patch embeddings as temporal windows for zero-transfer. ViTime~\cite{ViTime} operates in a binary image space with a ViT-based tokenizer and decoder, emphasizing periodicity and trend structures. DMMV~\cite{DMMV} adopts an MAE backbone as a visual forecaster, integrates it with a numerical forecaster, and enhances both via trend–seasonal decomposition.
Collectively, vision transformer-based backbones are favored for their ability to flexibly model both local and global dynamics.

\textbf{Vision–Language Backbone} Recent works explore vision–language models (VLMs) to enhance imaged time series modeling with multimodal alignment and interpretability. 
Time-VLM~\cite{Time-VLM} leverages pre-trained CLIP/BLIP/ViLT encoders to align line-plot images with textual descriptions, integrating temporal memory for forecasting. VLM-TSC~\cite{VLM-TSC} adapts LLaVA to jointly process imaged time series and textual prompts, enabling the language head to output class labels.
VLM4TS~\cite{VLM4TS} combines a CLIP vision encoder for anomaly candidate localization with GPT-4o for prompt-driven multimodal verification. These examples demonstrate how VLMs extend VFMs with semantic grounding and natural language reasoning. However, prompt design and alignment between modalities introduce new complexities.

\subsection{Tuning Strategy}
In terms of parameter updating during adaptation, the tuning strategies of VFMs largely mirror those of LFMs, encompassing tuning-free inference, promptless fine-tuning, and prompt-based fine-tuning. Both paradigms share the same objective—achieving efficient transfer under limited data and computational budgets. Nevertheless, differences in pre-training knowledge, model scale and  task formulation, lead to distinct emphases in how these strategies are applied. We summarize existing approaches into three groups and then discuss the differences of VFMs' tuning strategies to deepen understanding.

\textbf{Tuning-Free Inference} It refers to leveraging pre-trained VFMs directly on image based time series without updating model parameters, thereby achieving zero-shot generalization. VisionTS~\cite{VisionTS} reformulates forecasting as an image reconstruction task using MAE, enabling zero-shot prediction of future values through masked patch completion and inverse mapping. ITF-TAD~\cite{ITF-TAD} adopts a completely training-free approach by converting time series into Spectrogram and detecting anomalies in the feature space of a frozen ResNet foundation model with similarity-based scoring. VLM4TS~\cite{VLM4TS} employs a two-stage strategy where a pre-trained ViT localizes anomalies and a frozen GPT-4o verifies them through visual–textual reasoning, both stages requiring no parameter updates. These works highlight that tuning-free inference is particularly attractive in VFM-based adaptation, as pre-trained visual representations already capture structural patterns and allow effective transfer with minimal overhead.

\textbf{Promptless Fine-Tuning} It adapts VFMs to time series tasks by updating lightweight modules or task-specific heads, while freezing the majority of backbone parameters and without relying on prompts.
ViTST~\cite{ViTST} fine-tunes a pre-trained vision
transformer for irregularly sampled time series classification.
Besides zero-shot learning, VisionTS~\cite{VisionTS} also tests parameter-efficient fine-tuning of different modules and finds that only fine-tuning layer normalization is the best strategy.
ViTime~\cite{ViTime} introduces a visual tokenizer, decoder, and refining module to operate in a binary image space, requiring supervised training of these lightweight components while largely reusing pre-trained weights. 
DMMV~\cite{DMMV} freezes the majority of a visual backbone and only fine-tunes its normalization layer and a numerical forecaster for forecasting. 
Collectively, these approaches emphasize that promptless fine-tuning in VFMs often centers on adapters, normalization layers, or lightweight decoders, striking a balance between leveraging pre-trained knowledge and adapting to time series tasks.

\textbf{Prompt-based Fine-Tuning} It relies on prompts to guide adaptation, combined with parameter updates (e.g., tuning a vision–language projector).
Note that stand-alone VFMs (e.g., ViT~\cite{ViT}, MAE~\cite{MAE}) do not accept natural language prompts directly. Instead, prompt-based tuning typically involves vision-Language multimodal models (e.g., CLIP~\cite{CLIP}, LLaVA~\cite{LLaVA}), where the vision backbone is a VFM but prompts are injected through the language interface of a LFM. 
VLM-TSC~\cite{VLM-TSC} adapts LLaVA by mapping time series plots into the joint vision–language space and fine-tuning the vision–language projector so that the language head can output classification labels. 
Time-VLM~\cite{Time-VLM} leverages a vision–language framework, where the pre-trained visual and textual encoders are frozen, and the model is guided by multimodal prompts while fine-tuning lightweight temporal and fusion modules to produce forecasts.
These works illustrate that prompt-based fine-tuning for VFMs requires multimodal alignment: prompts are often jointly designed in both visual and textual modalities, enabling models to combine pre-trained vision knowledge with natural language reasoning for time series tasks.

\subsection{Task Configuration}
Compared to vision tasks where outputs are typically categorical labels or dense predictions over pixels, time series analysis requires models to produce numerical values aligned with specific tasks such as forecasting, imputation, generation and classification. Vision-based foundation models, therefore, need to redesign their output heads. This section summarizes how VFMs configure their task heads to produce the final outputs of different time series tasks.

\textbf{Classification Head} The simplest case is the classification head, where a fully connected or MLP layer maps output embeddings to class probabilities, supporting tasks such as classification and anomaly detection. 
ViTST~\cite{ViTST} follows this paradigm by converting irregularly sampled time series into line graph images and applying a vision transformer with a linear classifier to predict class labels. In the anomaly detection setting, ITF-TAD~\cite{ITF-TAD} replaces the standard classifier with a similarity-based scoring head over representations extracted from a frozen vision encoder.

\textbf{Image-based Prediction Head}
For numerical prediction tasks, the image-based head first predicts pixels of transformed time-series images, which are then mapped back to the original values.
VisionTS~\cite{VisionTS} treats forecasting as an image reconstruction task using MAE. The reconstructed image is then resized, de-normalized, and flattened back into a single-channel sequence to recover the forecasting window.
ViTime~\cite{ViTime} maps time series into binary images and utilizes a visual encoder to generate the predicted binary images, and then applies an inverse mapping to recover numerical values.
Similarly, DMMV~\cite{DMMV} performs forecasting by reconstructing masked regions of time-series images and then reversing them back into the temporal domain.

\textbf{Prompt-based Head}
In vision-language models, a prompt-based head allows models to generate task outputs in natural language form, thereby bypassing explicit task-specific heads. VLM4TS~\cite{VLM4TS} combines a lightweight vision-based anomaly screening stage (ViT4TS) with a verification stage where a VLM is prompted to refine anomaly intervals and provide natural-language justifications.


\section{Practical Applications}
\label{sec:applications}
\tikzstyle{leaf}=[draw=hiddendraw,
    rounded corners,
    minimum height=1em,
    text opacity=1, 
    align=center,
    fill opacity=.9,  
    text=black,
    align=left,
    font=\scriptsize,
    inner xsep=3pt,
    inner ysep=1pt,
    ]
\tikzstyle{middle}=[draw=hiddendraw,
    rounded corners,
    minimum height=1em,
    color = black,
    fill=leaf,
    text opacity=1, 
    align=center,
    text=black,
    align=center,
    font=\scriptsize,
    inner xsep=3pt,
    inner ysep=1pt,
    ]

\begin{figure}[t]
\centering
\begin{forest}
  for tree={
  forked edges,
  grow=east,
  reversed=true,
  anchor=base west,
  parent anchor=east,
  child anchor=west,
  base=middle,
  font=\scriptsize,
  rectangle,
  line width=0.7pt,
  edge={draw=black},
  draw=black,
  rounded corners,align=left,
  minimum width=2em,
    s sep=5pt,
    inner xsep=3pt,
    inner ysep=1pt},
  where level=1{text width=3em}{},
  where level=2{text width=17em,font=\scriptsize}{}, 
  where level=3{font=\scriptsize}{},
  where level=4{font=\scriptsize}{},
  where level=5{font=\scriptsize}{},
  [Applications, middle, rotate=90, anchor=north, draw=black, fill=main
        [Finance, middle, text width=3em, color=black, fill=middle, text=black
            [SSPT \cite{SSPT}{,} TDML \cite{TDML}{,} CAMEF \cite{CAMEF}{,} TWSN \cite{TWSN}{,} \\CIGN \cite{CIGN}{,} \cite{lopez2023can}{,} \cite{yu2023harnessing}{}
            ,leaf, text width=17em, color=black, fill=leaf, text=black]
        ]
        [Healthcare, middle, text width=3em, color=black, fill=middle, text=black
            [Brain-JEPA \cite{Brain-JEPA}{,} Brant-X \cite{Brant-X}{,} MERL \cite{MERL}{,} \\METS \cite{METS}{,}  MedTsLLM \cite{MedTsLLM}{,}  LLMFS \cite{LLMFS}{,}  Brant \cite{Brant} , leaf, text width=17em, color=black, fill=leaf, text=black]
        ]
        [Traffic, middle, text width=3em, color=black, fill=middle, text=black
            [UniST \cite{UniST}{,} UrbanGPT \cite{UrbanGPT}{,} GATGPT \cite{GATGPT}{,} \\STG-LLM \cite{STG-LLM}{,} FSTLLM \cite{FSTLLM}{,} ST-LLM \cite{ST-LLM}{,} \\LLMST \cite{LLMST}{,} LLM-Mob \cite{LLM-Mob}{,} AuxMobLCast \cite{AuxMobLCast}, leaf, text width=17em, color=black, fill=leaf, text=black]
        ]
        [Others, middle, text width=3em, color=black, fill=middle, text=black
            [ClimaX \cite{ClimaX}{,} BearingFM \cite{BearingFM}{,} UniMTS \cite{UniMTS}, leaf, text width=17em, color=black, fill=leaf, text=black]
        ]
    ]           
\end{forest}
\vspace{-5mm}
\caption{A domain taxonomy of foundation models for time series analysis.}
\label{fig:taxonomy}
\vspace{-4mm}
\end{figure}
This section focuses on the real-world applications of foundation models in time series analysis. We summarize their deployment of different domains such as finance, healthcare, and transportation. We highlight how time-series foundation models tackle domain-specific challenges, illustrating their adaptability and effectiveness in diverse real-world scenarios. 

\subsection{Finance}
Financial time series, such as stock prices, trading volumes, and macroeconomic indicators, are typically non-stationary, exhibit complex dependencies, and are highly sensitive to external events. These characteristics pose significant challenges for traditional statistical models and task-specific deep learning approaches. Foundation models (FMs), with their large-scale pre-training and strong transferability, offer a promising framework to capture  complex temporal dependencies, integrate contextual signals, and generalize across heterogeneous financial environments. Recent studies have begun exploring this promise. One line of works applies large language models (LLMs) for financial time series analysis. For instance, CIGN~\cite{CIGN} leverages ChatGPT to extract latent inter-company relations from financial news and subsequently constructs graphs for stock movement prediction. TWSN~\cite{TWSN} designs a multi-modal prompt that integrates historical stock price features with social media tweets, thereby eliciting ChatGPT’s reasoning ability for stock prediction tasks. TDML~\cite{TDML} further investigates the zero-shot inference capability of ChatGPT and GPT-4 in multimodal stock movement prediction. CAMEF~\cite{CAMEF} introduces a causal-augmented multimodality framework that links macroeconomic announcements with financial series, augmented by LLM-based counterfactual reasoning. 
Another line of research explores TSFM pre-trained on financial time series. For example, SSPT~\cite{SSPT} designs specialized pre-training tasks tailored to stock characteristics, such as sector classification and moving average prediction, demonstrating consistent improvements in investment return and Sharpe ratio. It underscores the value of tailoring TSFM directly to financial time series characteristics.
Overall, these studies demonstrate the potential of foundation models in finance, while emphasizing the importance of multimodal integration, domain-specific pre-training, and interpretability for practical deployment.

\subsection{Healthcare}
Healthcare is a key domain for applying foundation models (FMs) to time series, owing to the prevalence of physiological signals such as electrocardiograms (ECG), electroencephalograms (EEG), and intracranial recordings. These data are high-dimensional, noisy, and heterogeneous, making generalization and transfer learning particularly important. Unlike task-specific models trained on narrow datasets, FMs offer the promise of learning universal physiological representations that adapt across diverse clinical scenarios.
Recent works highlight this trend. LLMF~\cite{LLMFS} examines the current LLM's performance on health tasks with a limited number of training examples. They demonstrate that the PaLM with 24 billion parameters can digest time-series health data under few-shot learning setting and make meaningful inferences. METS~\cite{METS} leverages ClinicalBERT to align ECG signals with textual reports via contrastive learning, enhancing multimodal ECG classification.
MedTsLLM~\cite{MedTsLLM} integrates time series with textual context, enabling the model to leverage LLM reasoning for tasks such as segmentation, boundary detection, and anomaly detection.
MERL~\cite{MERL} explores multimodal ECG learning with clinical reports and LLM-generated knowledge-enhanced prompts, enabling zero-shot ECG classification.
Brant~\cite{Brant} proposes the largest FM pre-trained on a large corpus of intracranial recordings, capturing long-term temporal and frequency dependencies to support tasks such as seizure detection and neural signal forecasting.
Brant-X~\cite{Brant-X} extends this by aligning EEG with other physiological signals through a unified cross-signal framework.
These studies underscore FMs in healthcare time series, advancing multimodal integration, improving generalization under data scarcity, and enabling clinically relevant zero-shot applications.

\subsection{Traffic}
In intelligent transportation systems, traffic applications are fundamentally driven by spatio-temporal data, which are used to forecast traffic flows, estimate travel demand, and model mobility patterns. Unlike domains enriched with textual data, traffic tasks rely predominantly on sensor-based time series, which makes them more challenging for foundation models to model complex spatial–temporal dependencies.
Early explorations primarily focused on applying LLMs to specific mobility tasks. For example, LLM-Mob~\cite{LLM-Mob} employs GPT-3.5 with instruction prompts to capture both short- and long-term dependencies in trajectories, while LLMST~\cite{LLMST} leverages GPT-3.5/4 for anomaly detection by probing trajectory indicators and interactions. Moving beyond zero-shot inference, AuxMobLCast~\cite{AuxMobLCast} fine-tunes a BERT–GPT-2 encoder–decoder augmented with an auxiliary POI module to enhance trajectory forecasting.
Recent studies, in contrast, aim to establish more universal spatio-temporal foundation models. UniST~\cite{UniST} introduces a general pre-trained model for diverse urban tasks through knowledge-guided prompts, and UrbanGPT~\cite{UrbanGPT} combines a spatio-temporal dependency encoder with instruction tuning to improve generalization under data scarcity. Structural modeling has also been explored: GATGPT~\cite{GATGPT} integrates graph attention with LLMs for efficient spatio-temporal imputation, while STG-LLM~\cite{STG-LLM} designs a tokenizer–adapter framework to transform graph-structured data into LLM-friendly tokens. To improve robustness, FSTLLM~\cite{FSTLLM} investigates few-shot prediction with LLM priors, and ST-LLM~\cite{ST-LLM} incorporates spatio-temporal embeddings and partially frozen attention to capture global traffic flow dependencies.

\subsection{Other Domains}
Time series serve as the core data modality in diverse domains.
Beyond finance, traffic, and healthcare, foundation models for time series have also advanced in other domains, reflecting their broad applicability. In climate science, ClimaX~\cite{ClimaX} pre-trains on CMIP6 climate datasets with a self-supervised objective and demonstrates strong generalization across weather and climate tasks. In industrial applications, BearingFM~\cite{BearingFM} introduces a cloud–edge collaborative framework for bearing fault diagnosis, where cloud-based pre-training is combined with domain-specific fine-tuning at the edge, showing how foundation models can be adapted to resource-constrained scenarios. In human activity analysis, UniMTS~\cite{UniMTS} presents the first unified pre-training framework for motion time series from mobile and wearable devices, aligning time series with text descriptions via contrastive learning and achieving robustness to device heterogeneity and orientation shifts.
These efforts highlight the versatility of foundation models in different domains, underscoring their potential to unify diverse time series tasks within a generalizable framework.

\section{Future Directions}
This survey offers a comprehensive review of recent advances in foundation models for time series analysis. Despite the rapid progress, critical challenges persist, presenting opportunities for further breakthroughs. The  section highlights key future directions in this evolving field.

\label{sec:future}
\textbf{Backbone of Time Series-based Foundation Model}
Transformers remain the dominant backbone for time series foundation models owing to their strong sequence modeling and scalability. Yet, they struggle to capture intrinsic temporal features and incur high computational costs. Advanced variants such as Transformer++~\cite{TSsur-tran2022} attempt to alleviate these issues, while efficient alternatives like TTM~\cite{TTM}, built on a TSMixer backbone, highlight the potential of lightweight designs. State-Space Models (SSMs)~\cite{sarrof2024expressive} further provide promising directions with linear complexity and continuous-time dynamics. Looking ahead, emerging paradigms—such as diffusion models for uncertainty-aware generation~\cite{TimeDiT} and mixture-of-experts architectures for scalable adaptation~\cite{Time-MoE}—suggest new opportunities. Ultimately, hybrid approaches that integrate the strengths of different architectures may become central to advancing both the efficiency and effectiveness of time series foundation models.

\textbf{Time Series based Foundation Model for Classification}
Time series classification is fundamental to various real-world applications~\cite{DLsur-TS2019}. While as we can see in Table \ref{tab:tab_III_2}, most TSFMs focus on time series forecasting and only a few multi-task TSFMs incorporate the classification setting~\cite{UNITS}.
However, they face limitations including operating in a univariate setting or lacking zero-shot capabilities. Building a robust foundation model for classification requires overcoming several challenges, including inconsistent input lengths, varying numbers of variables, mismatched label spaces, and negative transfer across datasets. Additionally, widely used classification benchmarks (e.g., UCR~\cite{TSdat-UCR}, UEA~\cite{UEA}) remain relatively small, limiting model generalization. Future research should integrate larger classification datasets, such as diverse physiological and clinical datasets (e.g., those from PhysioNet). Moreover, future research should explore cross-task transfer between forecasting and classification, and investigate label-efficient strategies to reduce reliance on scarce labeled data.

\textbf{Interpretability}
Despite advances in LFM's adaptation, most efforts prioritize modality alignment and tuning strategy, overlooking interpretability.
However, in high-stakes domains  such as autonomous driving and healthcare, a lack of explainability hinders the real-world deployment of models.
Some studies have taken initial steps to improve LFM's interpretability. FPT~\cite{FPT} shows that self-attention of LFM behaves like principal component analysis, explaining its ability to bridge domain gaps. TEMPO~\cite{TEMPO} introduces an interpretable additive model to reveal how input components influence predictions.
Future research should prioritize: (1) Enhancing prompting techniques for better explanations. (2) Integrating interpretable modules into LFMs for time series tasks.
(3) Adopting post-hoc methods (e.g., LIME~\cite{LIME}) to improve decision transparency. These efforts strike a balance between performance and interpretability, leading to safer and more reliable LFMs for time series.

\textbf{Time Series Agents}
With the continuous advancement of agent technologies, their integration with time series analysis presents exciting opportunities for future research. Early explorations, such as Self-Extend-Agentic-RAG~\cite{SelfExtend-Agentic-RAG}, ~\cite{From-News}, ~\cite{SocioDojo} and TimeCAP~\cite{TimeCAP}, demonstrate that agents can coordinate sub-tasks or combine textual and temporal signals to address diverse challenges including forecasting, classification, imputation, and anomaly detection. Beyond task-oriented applications, frameworks like BRIDGE~\cite{BRIDGE} and TESSA~\cite{TESSA} highlight the potential of agents to generate and annotate time series data, thereby alleviating data scarcity and improving label quality.
Looking ahead, several promising directions emerge. First, developing agentic foundation models for time series could enable flexible orchestration across heterogeneous tasks, with agents dynamically adapting instruction-tuned modules. Second, multi-modal integration remains underexplored: leveraging agents to fuse textual knowledge, real-time signals, and structured data may substantially enhance robustness in high-stakes domains such as finance and healthcare. Third, data-centric agent systems that generate, curate, and annotate large-scale time series datasets could accelerate progress in representation learning. Finally, establishing systematic evaluation protocols for reasoning reliability, interpretability, and long-term decision-making will be crucial for deploying agent-based approaches in real-world scenarios.

\textbf{Time Series Reasoning}
Time series reasoning with large language models is emerging as a promising direction, moving beyond pattern recognition toward causal inference and decision-making. Real-world applications such as clinical diagnosis, financial risk assessment, and event forecasting demand reasoning over causal dependencies, temporal logic, and inter-variable relationships. Yet, its definition remains unsettled: ~\cite{ma2024language} frames time series reasoning as etiological reasoning, question answering, and context-aided forecasting, whereas~\cite{TTRL} describes it as a pipeline of perception, contextualization, and deductive reasoning. To evaluate this capability,~\cite{ma2024language} introduces a benchmark dataset and finds that LFMs, despite strong performance in other domains, exhibit limited ability on time series reasoning. In contrast,~\cite{ChatTS} proposes a multi-modal time-series LFM that integrates a lightweight encoder with chain-of-thought augmentation, enabling reasoning path generation and outperforming GPT-4o on zero-shot tasks. ~\cite{zhou2024are} further emphasizes the role of time series reasoning to bridge language and temporal data. 
Moving forward, advancing time series reasoning will require unified task definitions that clearly delineate its scope, along with standardized benchmarks and rigorous evaluation protocols to measure progress. Future research should also explore how to embed causal inference, leverage multimodal context, and produce interpretable reasoning paths. Such efforts will be critical for enabling LFMs to deliver robust, trustworthy, and human-like reasoning in high-stakes applications such as clinical diagnosis and financial risk assessment.

\textbf{Vision-based Foundation Models for Time Series Analysis}
Leveraging large vision models and vision-language models for time series is an emerging frontier, and current explorations remain limited compared with the surge of LFM-based methods.
By transforming time series into images (e.g., spectrograms or recurrence plots), recent studies show that pre-trained VFMs can be directly adapted: VisionTS~\cite{VisionTS} demonstrates strong zero-shot forecasting via visual masked autoencoders, while ViTime~\cite{ViTime} and ITF-TAD~\cite{ITF-TAD} explore vision transformers for forecasting and anomaly detection. Beyond pure vision, multimodal approaches such as Time-VLM~\cite{Time-VLM} enhance reasoning and interpretability in tasks like classification and anomaly detection.
Future research should explore more advanced imaging strategies to better preserve inter-variate correlations, investigate multi-view or decomposition-based representations, and extend vision-based models toward multimodal time series reasoning. These directions highlight the potential of VFMs for time series by leveraging visual priors and enabling richer cross-modal learning.

\section{Conclusion}
\label{sec:conclusion}
In this survey, we provide a comprehensive and up-to-date overview of foundation models for time series analysis. We introduce a modality-aware challenge taxonomy that highlights the distinct obstacles faced by time series–, language–, and vision–based foundation models, and we analyze and classify corresponding solutions within this framework. We also review representative applications across diverse domains, provide code resources to support reproducibility, and discuss future research opportunities. 
This survey provides insights into the progress achieved, the gaps that remain, and the directions where future breakthroughs are critical.
\bibliographystyle{IEEEtran}
\bibliography{IEEEabrv, ref}

\begin{thebibliography}{100}
\providecommand{\url}[1]{#1}
\csname url@samestyle\endcsname
\providecommand{\newblock}{\relax}
\providecommand{\bibinfo}[2]{#2}
\providecommand{\BIBentrySTDinterwordspacing}{\spaceskip=0pt\relax}
\providecommand{\BIBentryALTinterwordstretchfactor}{4}
\providecommand{\BIBentryALTinterwordspacing}{\spaceskip=\fontdimen2\font plus
\BIBentryALTinterwordstretchfactor\fontdimen3\font minus \fontdimen4\font\relax}
\providecommand{\BIBforeignlanguage}[2]{{%
\expandafter\ifx\csname l@#1\endcsname\relax
\typeout{** WARNING: IEEEtran.bst: No hyphenation pattern has been}%
\typeout{** loaded for the language `#1'. Using the pattern for}%
\typeout{** the default language instead.}%
\else
\language=\csname l@#1\endcsname
\fi
#2}}
\providecommand{\BIBdecl}{\relax}
\BIBdecl

\bibitem{SM-Mark1989}
J.~D. Hamilton, ``A new approach to the economic analysis of nonstationary time series and the business cycle,'' \emph{Econometrica: Journal of the econometric society}, pp. 357--384, 1989.

\bibitem{kontopoulou2023review}
V.~I. Kontopoulou, A.~D. Panagopoulos, I.~Kakkos, and G.~K. Matsopoulos, ``A review of arima vs. machine learning approaches for time series forecasting in data driven networks,'' \emph{Future Internet}, vol.~15, no.~8, p. 255, 2023.

\bibitem{SM-ARCH1982}
R.~F. Engle, ``Autoregressive conditional heteroscedasticity with estimates of the variance of united kingdom inflation,'' \emph{Econometrica: Journal of the econometric society}, pp. 987--1007, 1982.

\bibitem{al2023unique}
F.~S. Al-Duais and R.~S. Al-Sharpi, ``A unique markov chain monte carlo method for forecasting wind power utilizing time series model,'' \emph{Alexandria Engineering Journal}, vol.~74, pp. 51--63, 2023.

\bibitem{Transur-2018}
L.~Zhu, F.~R. Yu, Y.~Wang, B.~Ning, and T.~Tang, ``Big data analytics in intelligent transportation systems: A survey,'' \emph{IEEE Transactions on Intelligent Transportation Systems}, vol.~20, no.~1, pp. 383--398, 2018.

\bibitem{IOTsur-2018}
M.~Ge, H.~Bangui, and B.~Buhnova, ``Big data for internet of things: a survey,'' \emph{Future generation computer systems}, vol.~87, pp. 601--614, 2018.

\bibitem{SM-Ecommerce2016}
S.~Akter and S.~F. Wamba, ``Big data analytics in e-commerce: a systematic review and agenda for future research,'' \emph{Electronic Markets}, vol.~26, no.~2, pp. 173--194, 2016.

\bibitem{kong2025deep}
X.~Kong, Z.~Chen, W.~Liu, K.~Ning, L.~Zhang, S.~Muhammad~Marier, Y.~Liu, Y.~Chen, and F.~Xia, ``Deep learning for time series forecasting: a survey,'' \emph{International Journal of Machine Learning and Cybernetics}, pp. 1--34, 2025.

\bibitem{TS-CNNC}
B.~Zhao, H.~Lu, S.~Chen, J.~Liu, and D.~Wu, ``Convolutional neural networks for time series classification,'' \emph{Journal of Systems Engineering and Electronics}, vol.~28, no.~1, pp. 162--169, 2017.

\bibitem{TS-CNNF}
J.-F. Chen, W.-L. Chen, C.-P. Huang, S.-H. Huang, and A.-P. Chen, ``Financial time-series data analysis using deep convolutional neural networks,'' in \emph{2016 7th International conference on cloud computing and big data (CCBD)}, 2016, pp. 87--92.

\bibitem{TSsur-graph2023}
H.~Chen and H.~Eldardiry, ``Graph time-series modeling in deep learning: a survey,'' \emph{ACM Transactions on Knowledge Discovery from Data}, vol.~18, no.~5, pp. 1--35, 2024.

\bibitem{TS-graphF}
D.~Cheng, F.~Yang, S.~Xiang, and J.~Liu, ``Financial time series forecasting with multi-modality graph neural network,'' \emph{Pattern Recognition}, vol. 121, p. 108218, 2022.

\bibitem{TSsur-tran2022}
Q.~Wen, T.~Zhou, C.~Zhang, W.~Chen, Z.~Ma, J.~Yan, and L.~Sun, ``Transformers in time series: A survey,'' \emph{arXiv preprint arXiv:2202.07125}, 2022.

\bibitem{TSsur-dif2023}
L.~Lin, Z.~Li, R.~Li, X.~Li, and J.~Gao, ``Diffusion models for time-series applications: a survey,'' \emph{Frontiers of Information Technology \& Electronic Engineering}, vol.~25, no.~1, pp. 19--41, 2024.

\bibitem{METS}
J.~Li, C.~Liu, S.~Cheng, R.~Arcucci, and S.~Hong, ``Frozen language model helps ecg zero-shot learning,'' in \emph{Medical Imaging with Deep Learning (MIDL)}, vol. 227, 2024, pp. 402--415.

\bibitem{kim2021reversible}
T.~Kim, J.~Kim, Y.~Tae, C.~Park, J.-H. Choi, and J.~Choo, ``Reversible instance normalization for accurate time-series forecasting against distribution shift,'' in \emph{International conference on learning representations}, 2021.

\bibitem{sawhney2021fast}
R.~Sawhney, A.~Wadhwa, S.~Agarwal, and R.~Shah, ``Fast: Financial news and tweet based time aware network for stock trading,'' in \emph{Proceedings of the 16th conference of the european chapter of the association for computational linguistics: main volume}, 2021, pp. 2164--2175.

\bibitem{bommasani2021opportunities}
R.~Bommasani, D.~A. Hudson, E.~Adeli, R.~Altman, S.~Arora, S.~von Arx, M.~S. Bernstein, J.~Bohg, A.~Bosselut, E.~Brunskill \emph{et~al.}, ``On the opportunities and risks of foundation models,'' \emph{arXiv preprint arXiv:2108.07258}, 2021.

\bibitem{BERT}
J.~Devlin, M.-W. Chang, K.~Lee, and K.~Toutanova, ``Bert: Pre-training of deep bidirectional transformers for language understanding,'' in \emph{Proceedings of the 2019 conference of the North American chapter of the association for computational linguistics: human language technologies, volume 1 (long and short papers)}, 2019, pp. 4171--4186.

\bibitem{T5}
C.~Raffel, N.~Shazeer, A.~Roberts, K.~Lee, S.~Narang, M.~Matena, Y.~Zhou, W.~Li, and P.~J. Liu, ``Exploring the limits of transfer learning with a unified text-to-text transformer,'' \emph{The Journal of Machine Learning Research}, vol.~21, no.~1, pp. 5485--5551, 2020.

\bibitem{GPT-2}
A.~Radford, J.~Wu, R.~Child, D.~Luan, D.~Amodei, I.~Sutskever \emph{et~al.}, ``Language models are unsupervised multitask learners,'' \emph{OpenAI blog}, vol.~1, no.~8, p.~9, 2019.

\bibitem{PaLM}
A.~Chowdhery, S.~Narang, J.~Devlin, M.~Bosma, G.~Mishra, A.~Roberts, P.~Barham, H.~W. Chung, C.~Sutton, S.~Gehrmann \emph{et~al.}, ``Palm: Scaling language modeling with pathways,'' \emph{Journal of Machine Learning Research}, vol.~24, no. 240, pp. 1--113, 2023.

\bibitem{LLaMA}
H.~Touvron, T.~Lavril, G.~Izacard, X.~Martinet, M.-A. Lachaux, T.~Lacroix, B.~Rozi{\`e}re, N.~Goyal, E.~Hambro, F.~Azhar \emph{et~al.}, ``Llama: Open and efficient foundation language models,'' \emph{arXiv preprint arXiv:2302.13971}, 2023.

\bibitem{wei2022emergent}
J.~Wei, Y.~Tay, R.~Bommasani, C.~Raffel, B.~Zoph, S.~Borgeaud, D.~Yogatama, M.~Bosma, D.~Zhou, D.~Metzler \emph{et~al.}, ``Emergent abilities of large language models,'' \emph{Transactions on Machine Learning Research}, 2022.

\bibitem{GPT-3}
T.~B. Brown, B.~Mann, N.~Ryder, M.~Subbiah, J.~Kaplan, P.~Dhariwal, A.~Neelakantan, P.~Shyam, G.~Sastry, A.~Askell, S.~Agarwal, A.~Herbert-Voss, G.~Krueger, T.~Henighan, R.~Child, A.~Ramesh, D.~M. Ziegler, J.~Wu, C.~Winter, C.~Hesse, M.~Chen, E.~Sigler, M.~Litwin, S.~Gray, B.~Chess, J.~Clark, C.~Berner, S.~McCandlish, A.~Radford, I.~Sutskever, and D.~Amodei, ``Language models are few-shot learners,'' \emph{Advances in Neural Information Processing Systems}, vol.~33, pp. 1877--1901, 2020.

\bibitem{hendrycks2020measuring}
D.~Hendrycks, C.~Burns, S.~Basart, A.~Zou, M.~Mazeika, D.~Song, and J.~Steinhardt, ``Measuring massive multitask language understanding,'' in \emph{International Conference on Learning Representations}, 2020.

\bibitem{cot-2022}
J.~Wei, X.~Wang, D.~Schuurmans, M.~Bosma, F.~Xia, E.~Chi, Q.~V. Le, D.~Zhou \emph{et~al.}, ``Chain-of-thought prompting elicits reasoning in large language models,'' \emph{Advances in Neural Information Processing Systems}, vol.~35, pp. 24\,824--24\,837, 2022.

\bibitem{FPT}
T.~Zhou, P.~Niu, L.~Sun \emph{et~al.}, ``One fits all: Power general time series analysis by pretrained lm,'' in \emph{Advances in Neural Information Processing Systems}, vol.~36, 2023, pp. 43\,322--43\,355.

\bibitem{Time-LLM}
M.~Jin, S.~Wang, L.~Ma, Z.~Chu, J.~Zhang, X.~Shi, P.-Y. Chen, Y.~Liang, Y.-f. Li, S.~Pan \emph{et~al.}, ``Time-llm: Time series forecasting by reprogramming large language models,'' in \emph{International Conference on Learning Representations}, 2024.

\bibitem{TEMPO}
D.~Cao, F.~Jia, S.~O. Arik, T.~Pfister, Y.~Zheng, W.~Ye, and Y.~Liu, ``Tempo: Prompt-based generative pre-trained transformer for time series forecasting,'' in \emph{The Twelfth International Conference on Learning Representations}, 2024.

\bibitem{Moirai}
G.~Woo, C.~Liu, A.~Kumar \emph{et~al.}, ``Unified training of universal time series forecasting transformers,'' in \emph{Proceedings of the 41st International Conference on Machine Learning}, vol. 235, 2024, pp. 53\,140--53\,164.

\bibitem{Lag-Llama}
K.~Rasul, A.~Ashok, A.~R. Williams, A.~Khorasani, G.~Adamopoulos, R.~Bhagwatkar, M.~Bilo{\v{s}}, H.~Ghonia, N.~Hassen, A.~Schneider \emph{et~al.}, ``Lag-llama: Towards foundation models for time series forecasting,'' in \emph{R0-FoMo: Robustness of Few-shot and Zero-shot Learning in Large Foundation Models}, 2023.

\bibitem{Moment}
M.~Goswami, K.~Szafer, A.~Choudhry \emph{et~al.}, ``Moment: A family of open time-series foundation models,'' in \emph{Proceedings of the 41st International Conference on Machine Learning}, vol. 235, 2024, pp. 16\,115--16\,152.

\bibitem{zhou2023large}
M.~Jin, Q.~Wen, Y.~Liang, C.~Zhang, S.~Xue, X.~Wang, J.~Zhang, Y.~Wang, H.~Chen, X.~Li, S.~Pan, V.~S. Tseng, Y.~Zheng, L.~Chen, and H.~Xiong, ``Large models for time series and spatio-temporal data: A survey and outlook,'' \emph{arXiv:2310.10196}, 2023.

\bibitem{Miller2024survey}
J.~A. Miller, M.~Aldosari, F.~Saeed, N.~H. Barna, S.~Rana, I.~B. Arpinar, and N.~Liu, ``A survey of deep learning and foundation models for time series forecasting,'' \emph{arXiv preprint arXiv:2401.13912}, 2024.

\bibitem{jiang2024empowering}
Y.~Jiang, Z.~Pan, X.~Zhang, S.~Garg, A.~Schneider, Y.~Nevmyvaka, and D.~Song, ``Empowering time series analysis with large language models: a survey,'' in \emph{Proceedings of the Thirty-Third International Joint Conference on Artificial Intelligence}, 2024, pp. 8095--8103.

\bibitem{zhang2024large}
X.~Zhang, R.~R. Chowdhury, R.~K. Gupta, and J.~Shang, ``Large language models for time series: a survey,'' in \emph{Proceedings of the Thirty-Third International Joint Conference on Artificial Intelligence}, 2024, pp. 8335--8343.

\bibitem{jin2024position}
M.~Jin, Y.~Zhang, W.~Chen, K.~Zhang, Y.~Liang, B.~Yang, J.~Wang, S.~Pan, and Q.~Wen, ``Position: What can large language models tell us about time series analysis,'' in \emph{Proceedings of the 41st International Conference on Machine Learning}, 2024, pp. 22\,260--22\,276.

\bibitem{liang2024foundation}
Y.~Liang, H.~Wen, Y.~Nie, Y.~Jiang, M.~Jin, D.~Song, S.~Pan, and Q.~Wen, ``Foundation models for time series analysis: A tutorial and survey,'' in \emph{Proceedings of the ACM SIGKDD International Conference on Knowledge Discovery and Data Mining (KDD)}, 2024, pp. 4460--4470.

\bibitem{kottapalli2025foundation}
S.~R.~K. Kottapalli, K.~Hubli, S.~Chandrashekhara, G.~Jain, S.~Hubli, G.~Botla, and R.~Doddaiah, ``Foundation models for time series: A survey,'' \emph{arXiv preprint arXiv:2504.04011}, 2025.

\bibitem{ViT}
A.~Dosovitskiy, L.~Beyer, A.~Kolesnikov, D.~Weissenborn, X.~Zhai, T.~Unterthiner, M.~Dehghani, M.~Minderer, G.~Heigold, S.~Gelly \emph{et~al.}, ``An image is worth 16x16 words: Transformers for image recognition at scale,'' in \emph{International Conference on Learning Representations}, 2020.

\bibitem{MAE}
K.~He, X.~Chen, S.~Xie, Y.~Li, P.~Dollar, and R.~Girshick, ``Masked autoencoders are scalable vision learners,'' in \emph{Proceedings of the IEEE/CVF Conference on Computer Vision and Pattern Recognition (CVPR)}, 2022.

\bibitem{BEiT}
H.~Bao, L.~Dong, S.~Piao, and F.~Wei, ``Beit: Bert pre-training of image transformers,'' in \emph{International Conference on Learning Representations}, 2021.

\bibitem{PatchTST}
Y.~Nie, N.~H. Nguyen, P.~Sinthong, and J.~Kalagnanam, ``A time series is worth 64 words: Long-term forecasting with transformers,'' in \emph{International Conference on Learning Representations}, 2023.

\bibitem{wu2022timesnet}
H.~Wu, T.~Hu, Y.~Liu, H.~Zhou, J.~Wang, and M.~Long, ``Timesnet: Temporal 2d-variation modeling for general time series analysis,'' \emph{arXiv preprint arXiv:2210.02186}, 2022.

\bibitem{GPT-4}
J.~Achiam, S.~Adler, S.~Agarwal, L.~Ahmad, I.~Akkaya, F.~L. Aleman, D.~Almeida, J.~Altenschmidt, S.~Altman, S.~Anadkat \emph{et~al.}, ``Gpt-4 technical report,'' \emph{arXiv preprint arXiv:2303.08774}, 2023.

\bibitem{dubey2024llama}
A.~Dubey, A.~Jauhri, A.~Pandey, A.~Kadian, A.~Al-Dahle, A.~Letman, A.~Mathur, A.~Schelten, A.~Yang, A.~Fan \emph{et~al.}, ``The llama 3 herd of models,'' \emph{arXiv e-prints}, pp. arXiv--2407, 2024.

\bibitem{SimCLR}
T.~Chen, S.~Kornblith, M.~Norouzi, and G.~Hinton, ``A simple framework for contrastive learning of visual representations,'' in \emph{Proceedings of the 37th International Conference on Machine Learning (ICML)}, vol. 119, 2020, pp. 1597--1607.

\bibitem{CLIP}
A.~Radford, J.~W. Kim, C.~Hallacy, A.~Ramesh, G.~Goh, S.~Agarwal, G.~Sastry, A.~Askell, P.~Mishkin, J.~Clark, G.~Krueger, and I.~Sutskever, ``Learning transferable visual models from natural language supervision,'' in \emph{Proceedings of the 38th International Conference on Machine Learning (ICML)}, ser. Proceedings of Machine Learning Research, vol. 139, 2021, pp. 8748--8763.

\bibitem{ViLT}
W.~Kim, B.~Son, and I.~Kim, ``Vilt: Vision-and-language transformer without convolution or region supervision,'' in \emph{International conference on machine learning}, 2021, pp. 5583--5594.

\bibitem{BLIP}
J.~Li, D.~Li, C.~Xiong, and S.~Hoi, ``Blip: Bootstrapping language-image pre-training for unified vision-language understanding and generation,'' in \emph{International conference on machine learning}.\hskip 1em plus 0.5em minus 0.4em\relax PMLR, 2022, pp. 12\,888--12\,900.

\bibitem{LLaVA}
H.~Liu, C.~Li, Q.~Wu, and Y.~J. Lee, ``Visual instruction tuning,'' \emph{Advances in neural information processing systems}, vol.~36, pp. 34\,892--34\,916, 2023.

\bibitem{Kosmos-1}
S.~Huang, L.~Dong, W.~Wang, Y.~Hao, S.~Singhal, S.~Ma, T.~Lv, L.~Cui, O.~K. Mohammed, B.~Patra \emph{et~al.}, ``Language is not all you need: Aligning perception with language models,'' \emph{Advances in Neural Information Processing Systems}, vol.~36, pp. 72\,096--72\,109, 2023.

\bibitem{chen2025florence}
J.~Chen, J.~Yang, H.~Wu, D.~Li, J.~Gao, T.~Zhou, and B.~Xiao, ``Florence-vl: Enhancing vision-language models with generative vision encoder and depth-breadth fusion,'' in \emph{Proceedings of the Computer Vision and Pattern Recognition Conference}, 2025, pp. 24\,928--24\,938.

\bibitem{bai2025qwen2}
S.~Bai, K.~Chen, X.~Liu, J.~Wang, W.~Ge, S.~Song, K.~Dang, P.~Wang, S.~Wang, J.~Tang \emph{et~al.}, ``Qwen2. 5-vl technical report,'' \emph{arXiv preprint arXiv:2502.13923}, 2025.

\bibitem{SSPT}
M.~Wang, T.~Ma, and S.~B. Cohen, ``Pre-training time series models with stock data customization,'' in \emph{Proceedings of the 31st ACM SIGKDD Conference on Knowledge Discovery and Data Mining V. 2}, 2025, pp. 3019--3030.

\bibitem{liu2024combining}
X.~Liu and Q.~Zhang, ``Combining seasonal and trend decomposition using loess with a gated recurrent unit for climate time series forecasting,'' \emph{IEEE Access}, vol.~12, pp. 85\,275--85\,290, 2024.

\bibitem{DAM}
L.~N. Darlow, Q.~Deng, A.~Hassan \emph{et~al.}, ``Dam: Towards a foundation model for forecasting,'' in \emph{Proceedings of the Twelfth International Conference on Learning Representations (ICLR)}, 2024.

\bibitem{GTT}
C.~Feng, L.~Huang, and D.~Krompass, ``General time transformer: an encoder-only foundation model for zero-shot multivariate time series forecasting,'' in \emph{Proceedings of the 33rd ACM International Conference on Information and Knowledge Management (CIKM)}, 2024, pp. 3757--3761.

\bibitem{TTM}
V.~Ekambaram, A.~Jati, P.~Dayama \emph{et~al.}, ``Tiny time mixers (ttms): Fast pre-trained models for enhanced zero/few-shot forecasting of multivariate time series,'' in \emph{Advances in Neural Information Processing Systems}, vol.~37, 2024, pp. 74\,147--74\,181.

\bibitem{TimeCAP}
G.~Lee, W.~Yu, K.~Shin, W.~Cheng, and H.~Chen, ``Timecap: Learning to contextualize, augment, and predict time series events with large language model agents,'' in \emph{Proceedings of the AAAI Conference on Artificial Intelligence}, vol.~39, no.~17, 2025, pp. 18\,082--18\,090.

\bibitem{ChatTime}
C.~Wang, Q.~Qi, J.~Wang, H.~Sun, Z.~Zhuang, J.~Wu, L.~Zhang, and J.~Liao, ``Chattime: A unified multimodal time series foundation model bridging numerical and textual data,'' in \emph{Proceedings of the AAAI Conference on Artificial Intelligence}, vol.~39, no.~12, 2025, pp. 12\,694--12\,702.

\bibitem{UNITS}
S.~Gao, T.~Koker, O.~Queen \emph{et~al.}, ``Units: A unified multi-task time series model,'' in \emph{Advances in Neural Information Processing Systems}, vol.~37, 2024, pp. 140\,589--140\,631.

\bibitem{Kim2022ReversibleIN}
T.~Kim, J.~Kim, Y.~Tae, C.~Park, J.~Choi, and J.~Choo, ``Reversible instance normalization for accurate time-series forecasting against distribution shift,'' in \emph{International Conference on Learning Representations}, 2022.

\bibitem{Unisur-2024}
P.~Trirat, Y.~Shin, J.~Kang, Y.~Nam, J.~Na, M.~Bae, J.~Kim, B.~Kim, and J.-G. Lee, ``Universal time-series representation learning: A survey,'' \emph{arXiv preprint arXiv:2401.03717}, 2024.

\bibitem{Timer}
Y.~Liu, H.~Zhang, C.~Li \emph{et~al.}, ``Timer: Generative pre-trained transformers are large time series models,'' in \emph{Proceedings of the 41st International Conference on Machine Learning}, vol. 235, 2024, pp. 32\,369--32\,399.

\bibitem{Time-MoE}
X.~Shi, S.~Wang, Y.~Nie, D.~Li, Z.~Ye, Q.~Wen, and M.~Jin, ``Time-moe: Billion-scale time series foundation models with mixture of experts,'' in \emph{International Conference on Learning Representations (ICLR)}, 2025.

\bibitem{Sundial}
Y.~Liu, G.~Qin, Z.~Shi, Z.~Chen, C.~Yang, X.~Huang, J.~Wang, and M.~Long, ``Sundial: A family of highly capable time series foundation models,'' in \emph{International Conference on Machine Learning (ICML)}, 2025.

\bibitem{TimeDiT}
D.~Cao, W.~Ye, Y.~Zhang, and Y.~Liu, ``Timedit: General-purpose diffusion transformers for time series foundation model,'' in \emph{International Conference on Machine Learning (ICML) Workshop on Foundation Models in the Wild}, 2024.

\bibitem{ForecastPFN}
S.~Dooley, G.~S. Khurana, C.~Mohapatra \emph{et~al.}, ``Forecastpfn: Synthetically-trained zero-shot forecasting,'' in \emph{Advances in Neural Information Processing Systems}, vol.~36, 2023, pp. 2403--2426.

\bibitem{TimeCLR}
C.~M. Yeh, X.~Dai, H.~Chen, Y.~Zheng, X.~Yan, W.~Zhang, S.~Wu, L.~Wang, and W.~Tu, ``Toward a foundation model for time series data,'' in \emph{Proceedings of the 32nd ACM International Conference on Information and Knowledge Management (CIKM)}, 2023, pp. 4400--4404.

\bibitem{FFTS}
S.~Chen, G.~Long, J.~Jiang \emph{et~al.}, ``Federated foundation models on heterogeneous time series,'' in \emph{Proceedings of the AAAI Conference on Artificial Intelligence}, vol.~39, no.~15, 2025, pp. 15\,839--15\,847.

\bibitem{LPTM}
P.~K. H and B.~A. Prakash, ``Large pre-trained time series models for cross-domain time series analysis tasks,'' in \emph{Advances in Neural Information Processing Systems}, vol.~37, 2024, pp. 56\,190--56\,214.

\bibitem{Chronos}
A.~F. Ansari, L.~Stella, C.~Turkmen, X.~Zhang, P.~Mercado, H.~Shen, O.~Shchur, S.~S. Rangapuram, S.~P. Arango, S.~Kapoor \emph{et~al.}, ``Chronos: Learning the language of time series,'' \emph{arXiv preprint arXiv:2403.07815}, 2024.

\bibitem{TOTEM}
S.~Talukder, Y.~Yue, and G.~Gkioxari, ``Totem: Tokenized time series embeddings for general time series analysis,'' \emph{arXiv preprint arXiv:2402.16412}, 2024.

\bibitem{TimeGPT}
A.~Garza, C.~Challu, and M.~Mergenthaler-Canseco, ``Timegpt-1,'' \emph{arXiv:2310.03589}, 2023.

\bibitem{TimeRAF}
H.~Zhang, C.~Xu, Y.-F. Zhang, Z.~Zhang, L.~Wang, and J.~Bian, ``Timeraf: Retrieval-augmented foundation model for zero-shot time series forecasting,'' \emph{IEEE Transactions on Knowledge \& Data Engineering}, no.~01, pp. 1--12, 2025.

\bibitem{ROSE}
Y.~Wang, Y.~Qiu, P.~Chen, K.~Zhao, Y.~Shu, Z.~Rao, L.~Pan, B.~Yang, and C.~Guo, ``Towards a general time series forecasting model with unified representation and adaptive transfer,'' in \emph{Forty-second International Conference on Machine Learning}, 2025.

\bibitem{MOIRAI-MOE}
X.~Liu, J.~Liu, G.~Woo, T.~Aksu, Y.~Liang, R.~Zimmermann, C.~Liu, S.~Savarese, C.~Xiong, and D.~Sahoo, ``Moirai-moe: Empowering time series foundation models with sparse mixture of experts,'' in \emph{International Conference on Machine Learning (ICML)}, 2025.

\bibitem{TimesFM}
A.~Das, W.~Kong, R.~Sen \emph{et~al.}, ``A decoder-only foundation model for time-series forecasting,'' in \emph{Proceedings of the 41st International Conference on Machine Learning}, vol. 235, 2024, pp. 10\,148--10\,167.

\bibitem{TimesFM-ICF}
A.~Das, M.~Faw \emph{et~al.}, ``In-context fine-tuning for time-series foundation models,'' in \emph{International Conference on Machine Learning (ICML)}, 2025.

\bibitem{TSdat-Monash}
R.~W. Godahewa, C.~Bergmeir, G.~I. Webb, R.~Hyndman, and P.~Montero-Manso, ``Monash time series forecasting archive,'' in \emph{Thirty-fifth Conference on Neural Information Processing Systems Datasets and Benchmarks Track (Round 2)}, 2021.

\bibitem{TSdat-UCI}
A.~Asuncion, ``Uci machine learning repository,'' 2007.

\bibitem{RevIN}
T.~Kim, J.~Kim, Y.~Tae, C.~Park, J.-H. Choi, and J.~Choo, ``Reversible instance normalization for accurate time-series forecasting against distribution shift,'' in \emph{International Conference on Learning Representations}, 2021.

\bibitem{UniTime}
X.~Liu, J.~Hu, Y.~Li \emph{et~al.}, ``Unitime: A language-empowered unified model for cross-domain time series forecasting,'' in \emph{Proceedings of the ACM Web Conference 2024}, 2024, pp. 4095--4106.

\bibitem{yao2025towards}
Q.~Yao, C.-H.~H. Yang, R.~Jiang, Y.~Liang, M.~Jin, and S.~Pan, ``Towards neural scaling laws for time series foundation models,'' in \emph{The Thirteenth International Conference on Learning Representations (ICLR 2025)}, 2025.

\bibitem{LLMTime}
N.~Gruver, M.~Finzi, S.~Qiu \emph{et~al.}, ``Large language models are zero-shot time series forecasters,'' in \emph{Advances in Neural Information Processing Systems}, vol.~36, 2023, pp. 19\,622--19\,635.

\bibitem{PromptCast}
H.~Xue and F.~D. Salim, ``Promptcast: A new prompt-based learning paradigm for time series forecasting,'' \emph{IEEE Transactions on Knowledge and Data Engineering}, vol.~36, no.~11, pp. 6851--6864, 2023.

\bibitem{SIGLLM}
S.~Alnegheimish, L.~Nguyen, L.~Berti-Equille, and K.~Veeramachaneni, ``Large language models can be zero-shot anomaly detectors for time series?'' \emph{arXiv preprint arXiv:2405.14755}, 2024.

\bibitem{TableTime}
J.~Wang, M.~Cheng, Q.~Mao, Q.~Liu, F.~Xu, X.~Li, and E.~Chen, ``Tabletime: Reformulating time series classification as zero-shot table understanding via large language models,'' \emph{arXiv e-prints}, pp. arXiv--2411, 2024.

\bibitem{LSTPrompt}
H.~Liu, Z.~Zhao, J.~Wang, H.~Kamarthi, and B.~A. Prakash, ``Lstprompt: Large language models as zero-shot time series forecasters by long-short-term prompting,'' \emph{arXiv:2402.16132}, 2024.

\bibitem{CALF}
P.~Liu, H.~Guo, T.~Dai \emph{et~al.}, ``Calf: Aligning llms for time series forecasting via cross-modal fine-tuning,'' in \emph{Proceedings of the AAAI Conference on Artificial Intelligence}, vol.~39, no.~18, 2025, pp. 18\,915--18\,923.

\bibitem{ISTS-PLM}
W.~Zhang, C.~Yin, H.~Liu, and H.~Xiong, ``Unleashing the power of pre-trained language models for irregularly sampled time series,'' in \emph{Proceedings of the 31st ACM SIGKDD Conference on Knowledge Discovery and Data Mining V. 2}, 2025, pp. 3831--3842.

\bibitem{LLM4TS}
C.~Chang, W.-C. Peng, and T.-F. Chen, ``Llm4ts: Two-stage fine-tuning for time-series forecasting with pre-trained llms,'' \emph{CoRR}, 2023.

\bibitem{aLLM4TS}
Y.~Bian, X.~Ju, J.~Li, Z.~Xu, D.~Cheng, and Q.~Xu, ``Multi-patch prediction: Adapting language models for time series representation learning,'' in \emph{Proceedings of the 41st International Conference on Machine Learning}, 2024, pp. 1--24.

\bibitem{AnomalyLLM}
C.~Liu, S.~He, Q.~Zhou \emph{et~al.}, ``Large language model guided knowledge distillation for time series anomaly detection,'' in \emph{Proceedings of the Thirty-Third International Joint Conference on Artificial Intelligence (IJCAI)}, 2024, pp. 2162--2170.

\bibitem{MedualTime}
Y.~Jiexia, W.~Zhang, Z.~Li, J.~Li, and F.~Tsung, ``Dualtime: A dual-adapter language model for time series multimodal representation learning,'' in \emph{International Joint Conference on Artificial Intelligence (IJCAI)}, 2025.

\bibitem{LeRet}
Q.~Huang, Z.~Zhou, K.~Yang \emph{et~al.}, ``Leret: Language-empowered retentive network for time series forecasting,'' in \emph{Proceedings of the Thirty-Third International Joint Conference on Artificial Intelligence}, 2024, pp. 4165--4173.

\bibitem{From-News}
X.~Wang, M.~Feng, J.~Qiu \emph{et~al.}, ``From news to forecast: Integrating event analysis in llm-based time series forecasting with reflection,'' in \emph{Advances in Neural Information Processing Systems}, vol.~37, 2024, pp. 58\,118--58\,153.

\bibitem{TEST}
C.~Sun, H.~Li, Y.~Li, and S.~Hong, ``Test: Text prototype aligned embedding to activate llm's ability for time series,'' in \emph{International Conference on Learning Representations (ICLR)}, 2024.

\bibitem{S2IP-LLM}
Z.~Pan, Y.~Jiang, S.~Garg, X.~Zhang, Y.~Nevmyvaka, and D.~Song, ``{$S^2$IP-LLM}: Semantic space informed prompt learning with {LLM} for time series forecasting,'' in \emph{Proceedings of the 41st International Conference on Machine Learning}, 2024, pp. 39\,135--39\,153.

\bibitem{NuwaTS}
J.~Cheng, C.~Yang, W.~Cai, Y.~Liang, Q.~Wen, and Y.~Wu, ``Nuwats: a foundation model mending every incomplete time series,'' \emph{arXiv preprint arXiv:2405.15317}, 2024.

\bibitem{Time-FFM}
Q.~Liu, X.~Liu, C.~Liu \emph{et~al.}, ``Time-ffm: Towards lm-empowered federated foundation model for time series forecasting,'' in \emph{Advances in Neural Information Processing Systems}, vol.~37, 2024, pp. 94\,512--94\,538.

\bibitem{AutoTimes}
Y.~Liu, G.~Qin, X.~Huang \emph{et~al.}, ``Autotimes: Autoregressive time series forecasters via large language models,'' in \emph{Advances in Neural Information Processing Systems}, vol.~37, 2024, pp. 122\,154--122\,184.

\bibitem{HiTime}
X.~Tao, T.~Pan, M.~Cheng, and Y.~Luo, ``Hierarchical multimodal llms with semantic space alignment for enhanced time series classification,'' \emph{arXiv:2410.18686}, 2024.

\bibitem{GPT4MTS}
F.~Jia, K.~Wang, Y.~Zheng \emph{et~al.}, ``Gpt4mts: Prompt-based large language model for multimodal time-series forecasting,'' in \emph{Proceedings of the AAAI Conference on Artificial Intelligence}, vol.~38, no.~21, 2024, pp. 23\,343--23\,351.

\bibitem{ExoLLM}
Q.~Huang, Z.~Zhou, K.~Yang \emph{et~al.}, ``Exploiting language power for time series forecasting with exogenous variables,'' in \emph{Proceedings of the ACM Web Conference 2025}, 2025, pp. 4043--4052.

\bibitem{InstructTime}
M.~Cheng, Y.~Chen, Q.~Liu \emph{et~al.}, ``Instructime: Advancing time series classification with multimodal language modeling,'' in \emph{Proceedings of the 18th ACM International Conference on Web Search and Data Mining (WSDM)}, 2025, pp. 792--800.

\bibitem{FSCA}
Y.~Hu, Q.~Li, D.~Zhang, J.~Yan, and Y.~Chen, ``Context-alignment: Activating and enhancing llms capabilities in time series,'' in \emph{International Conference on Learning Representations (ICLR)}, 2025.

\bibitem{STEM-LTS}
Z.~Zhao, P.~Wang, H.~Wen, S.~Wang, L.~Yu, and Y.~Wang, ``Stem-lts: Integrating semantic-temporal dynamics in llm-driven time series analysis,'' in \emph{Proceedings of the AAAI Conference on Artificial Intelligence}, vol.~39, no.~21, 2025, pp. 22\,858--22\,866.

\bibitem{LangTime}
W.~Niu, Z.~Xie, Y.~Sun, W.~He, M.~Xu, and C.~Hao, ``Langtime: A language-guided unified model for time series forecasting with proximal policy optimization,'' in \emph{International Conference on Machine Learning (ICML)}, 2025.

\bibitem{spathis2024first}
D.~Spathis and F.~Kawsar, ``The first step is the hardest: Pitfalls of representing and tokenizing temporal data for large language models,'' \emph{Journal of the American Medical Informatics Association}, vol.~31, no.~9, pp. 2151--2158, 2024.

\bibitem{LoRA}
E.~J. Hu, Y.~Shen, P.~Wallis, Z.~Allen-Zhu, Y.~Li, S.~Wang, L.~Wang, W.~Chen \emph{et~al.}, ``Lora: Low-rank adaptation of large language models.'' \emph{ICLR}, vol.~1, no.~2, p.~3, 2022.

\bibitem{ITF-TAD}
N.~Namura and Y.~Ichikawa, ``Training-free time-series anomaly detection: Leveraging image foundation models,'' \emph{arXiv preprint arXiv:2408.14756}, 2024.

\bibitem{VLM4TS}
Z.~He, S.~Alnegheimish, and M.~Reimherr, ``Harnessing vision-language models for time series anomaly detection,'' \emph{arXiv preprint arXiv:2506.06836}, 2025.

\bibitem{VisionTS}
M.~Chen, L.~Shen, Z.~Li, X.~J. Wang, J.~Sun, and C.~Liu, ``Visionts: Visual masked autoencoders are free-lunch zero-shot time series forecasters,'' in \emph{Proceedings of the 42nd International Conference on Machine Learning (ICML)}, 2025.

\bibitem{ViTST}
Z.~Li, S.~Li, and X.~Yan, ``Time series as images: Vision transformer for irregularly sampled time series,'' in \emph{Advances in Neural Information Processing Systems}, vol.~36, 2023, pp. 49\,187--49\,204.

\bibitem{ViTime}
L.~Yang, Y.~Wang, X.~Fan, I.~Cohen, J.~Chen, Y.~Zhao, and Z.~Zhang, ``Vitime: A visual intelligence-based foundation model for time series forecasting,'' \emph{arXiv preprint arXiv:2407.07311}, 2024.

\bibitem{DMMV}
C.~Shen, W.~Yu, Z.~Zhao, D.~Song, W.~Cheng, H.~Chen, and J.~Ni, ``Multi-modal view enhanced large vision models for long-term time series forecasting,'' \emph{arXiv preprint arXiv:2505.24003}, 2025.

\bibitem{VisionTS++}
L.~Shen, M.~Chen, X.~Liu, H.~Fu, X.~Ren, J.~Sun, Z.~Li, and C.~Liu, ``Visionts++: Cross-modal time series foundation model with continual pre-trained visual backbones,'' \emph{arXiv preprint arXiv:2508.04379}, 2025.

\bibitem{LDM4TS}
W.~Ruan, S.~Zhong, H.~Wen, and Y.~Liang, ``Vision-enhanced time series forecasting via latent diffusion models,'' \emph{arXiv preprint arXiv:2502.14887}, 2025.

\bibitem{OccamVTS}
S.~Lyu, S.~Zhong, W.~Ruan, Q.~Liu, Q.~Wen, H.~Xiong, and Y.~Liang, ``Occamvts: Distilling vision models to 1\% parameters for time series forecasting,'' \emph{arXiv preprint arXiv:2508.01727}, 2025.

\bibitem{VLM-TSC}
V.~Prithyani, M.~Mohammed, R.~Gadgil, R.~Buitrago, V.~Jain, and A.~Chadha, ``On the feasibility of vision-language models for time-series classification,'' \emph{arXiv preprint arXiv:2412.17304}, 2024.

\bibitem{Time-VLM}
S.~Zhong, W.~Ruan, M.~Jin, H.~Li, Q.~Wen, and Y.~Liang, ``Time-vlm: Exploring multimodal vision-language models for augmented time series forecasting,'' in \emph{International Conference on Machine Learning (ICML), Poster}, 2025.

\bibitem{TDML}
X.~Yu, Z.~Chen, and Y.~Lu, ``Harnessing llms for temporal data-a study on explainable financial time series forecasting,'' in \emph{Proceedings of the 2023 Conference on Empirical Methods in Natural Language Processing: Industry Track}, 2023, pp. 739--753.

\bibitem{CAMEF}
Y.~Zhang, W.~Yang, J.~Wang, Q.~Ma, and J.~Xiong, ``Camef: Causal-augmented multi-modality event-driven financial forecasting by integrating time series patterns and salient macroeconomic announcements,'' in \emph{Proceedings of the 31st ACM SIGKDD Conference on Knowledge Discovery and Data Mining V. 2}, 2025, pp. 3867--3878.

\bibitem{TWSN}
Q.~Xie, W.~Han, Y.~Lai, M.~Peng, and J.~Huang, ``The wall street neophyte: A zero-shot analysis of chatgpt over multimodal stock movement prediction challenges,'' \emph{arXiv preprint arXiv:2304.05351}, 2023.

\bibitem{CIGN}
Z.~Chen, L.~N. Zheng, C.~Lu, J.~Yuan, and D.~Zhu, ``Chatgpt informed graph neural network for stock movement prediction,'' \emph{arXiv preprint arXiv:2306.03763}, 2023.

\bibitem{lopez2023can}
A.~Lopez-Lira and Y.~Tang, ``Can chatgpt forecast stock price movements? return predictability and large language models,'' \emph{arXiv preprint arXiv:2304.07619}, 2023.

\bibitem{yu2023harnessing}
X.~Yu, Z.~Chen, and Y.~Lu, ``Harnessing llms for temporal data-a study on explainable financial time series forecasting,'' in \emph{Proceedings of the 2023 Conference on Empirical Methods in Natural Language Processing: Industry Track}, 2023, pp. 739--753.

\bibitem{Brain-JEPA}
Z.~Dong, R.~Li, Y.~Wu \emph{et~al.}, ``Brain-jepa: Brain dynamics foundation model with gradient positioning and spatiotemporal masking,'' in \emph{Advances in Neural Information Processing Systems}, vol.~37, 2024, pp. 86\,048--86\,073.

\bibitem{Brant-X}
D.~Zhang, Z.~Yuan, J.~Chen, K.~Chen, and Y.~Yang, ``Brant-x: A unified physiological signal alignment framework,'' in \emph{Proceedings of the 30th ACM SIGKDD Conference on Knowledge Discovery and Data Mining}, 2024, pp. 4155--4166.

\bibitem{MERL}
C.~Liu, Z.~Wan, C.~Ouyang, A.~Shah, W.~Bai, and R.~Arcucci, ``Zero-shot ecg classification with multimodal learning and test-time clinical knowledge enhancement,'' in \emph{Proceedings of the 41st International Conference on Machine Learning (ICML)}, vol. 235, 2024, pp. 31\,949--31\,963.

\bibitem{MedTsLLM}
N.~Chan, F.~Parker, W.~Bennett, T.~Wu, M.~Y. Jia, J.~Fackler, and K.~Ghobadi, ``Medtsllm: Leveraging llms for multimodal medical time series analysis,'' \emph{arXiv preprint arXiv:2408.07773}, 2024.

\bibitem{LLMFS}
X.~Liu, D.~McDuff, G.~Kovacs, I.~Galatzer-Levy, J.~Sunshine, J.~Zhan, M.-Z. Poh, S.~Liao, P.~Di~Achille, and S.~Patel, ``Large language models are few-shot health learners,'' \emph{arXiv:2305.15525}, 2023.

\bibitem{Brant}
D.~Zhang, Z.~Yuan, Y.~Yang \emph{et~al.}, ``Brant: Foundation model for intracranial neural signal,'' in \emph{Advances in Neural Information Processing Systems}, vol.~36, 2023, pp. 26\,304--26\,321.

\bibitem{UniST}
Y.~Yuan, J.~Ding, J.~Feng, D.~Jin, and Y.~Li, ``Unist: A prompt-empowered universal model for urban spatio-temporal prediction,'' in \emph{Proceedings of the 30th ACM SIGKDD Conference on Knowledge Discovery and Data Mining}, 2024, pp. 4095--4106.

\bibitem{UrbanGPT}
Z.~Li, L.~Xia, J.~Tang, Y.~Xu, L.~Shi, L.~Xia, D.~Yin, and C.~Huang, ``Urbangpt: Spatio-temporal large language models,'' in \emph{Proceedings of the 30th ACM SIGKDD Conference on Knowledge Discovery and Data Mining}, 2024, pp. 5351--5362.

\bibitem{GATGPT}
Y.~Chen, X.~Wang, and G.~Xu, ``Gatgpt: A pre-trained large language model with graph attention network for spatiotemporal imputation,'' \emph{arXiv:2311.14332}, 2023.

\bibitem{STG-LLM}
L.~Liu, S.~Yu, R.~Wang, Z.~Ma, and Y.~Shen, ``How can large language models understand spatial-temporal data?'' \emph{arXiv preprint arXiv:2401.14192}, 2024.

\bibitem{FSTLLM}
Y.~JIANG, Y.~Chen, X.~Li, Q.~Chao, S.~LIU, and G.~Cong, ``{FSTLLM}: Spatio-temporal {LLM} for few shot time series forecasting,'' in \emph{Forty-second International Conference on Machine Learning(ICML)}, 2025.

\bibitem{ST-LLM}
C.~Liu, S.~Yang, Q.~Xu, Z.~Li, C.~Long, Z.~Li, and R.~Zhao, ``Spatial-temporal large language model for traffic prediction,'' in \emph{2024 25th IEEE International Conference on Mobile Data Management (MDM)}, 2024, pp. 31--40.

\bibitem{LLMST}
Z.~Zhang, H.~Amiri, Z.~Liu, L.~Zhao, and A.~Z{\"u}fle, ``Large language models for spatial trajectory patterns mining,'' in \emph{Proceedings of the 1st ACM SIGSPATIAL International Workshop on Geospatial Anomaly Detection}, 2024, pp. 52--55.

\bibitem{LLM-Mob}
X.~Wang, M.~Fang, Z.~Zeng, and T.~Cheng, ``Where would i go next? large language models as human mobility predictors,'' \emph{arXiv preprint arXiv:2308.15197}, 2023.

\bibitem{AuxMobLCast}
H.~Xue, B.~P. Voutharoja, and F.~D. Salim, ``Leveraging language foundation models for human mobility forecasting,'' in \emph{Proceedings of the 30th international conference on advances in geographic information systems}, 2022, pp. 1--9.

\bibitem{ClimaX}
T.~Nguyen, J.~Brandstetter, A.~Kapoor \emph{et~al.}, ``Climax: A foundation model for weather and climate,'' in \emph{Proceedings of the 40th International Conference on Machine Learning}, 2023, pp. 25\,904--25\,938.

\bibitem{BearingFM}
Z.~Lai, C.~Yang, S.~Lan, L.~Wang, W.~Shen, and L.~Zhu, ``Bearingfm: Towards a foundation model for bearing fault diagnosis by domain knowledge and contrastive learning,'' \emph{International Journal of Production Economics}, vol. 275, p. 109319, 2024.

\bibitem{UniMTS}
X.~Zhang, D.~Teng, R.~R. Chowdhury \emph{et~al.}, ``Unimts: Unified pre-training for motion time series,'' in \emph{Advances in Neural Information Processing Systems}, vol.~37, 2024, pp. 107\,469--107\,493.

\bibitem{sarrof2024expressive}
Y.~Sarrof, Y.~Veitsman, and M.~Hahn, ``The expressive capacity of state space models: A formal language perspective,'' \emph{Advances in Neural Information Processing Systems}, vol.~37, pp. 41\,202--41\,241, 2024.

\bibitem{DLsur-TS2019}
H.~Ismail~Fawaz, G.~Forestier, J.~Weber, L.~Idoumghar, and P.-A. Muller, ``Deep learning for time series classification: a review,'' \emph{Data mining and knowledge discovery}, vol.~33, no.~4, pp. 917--963, 2019.

\bibitem{TSdat-UCR}
H.~A. Dau, A.~Bagnall, K.~Kamgar, C.-C.~M. Yeh, Y.~Zhu, S.~Gharghabi, C.~A. Ratanamahatana, and E.~Keogh, ``The ucr time series archive,'' \emph{IEEE/CAA Journal of Automatica Sinica}, p. 1293–1305, Nov 2019.

\bibitem{UEA}
A.~Bagnall, H.~A. Dau, J.~Lines, M.~Flynn, J.~Large, A.~Bostrom, P.~Southam, and E.~Keogh, ``The uea multivariate time series classification archive, 2018,'' \emph{arXiv preprint arXiv:1811.00075}, 2018.

\bibitem{LIME}
M.~T. Ribeiro, S.~Singh, and C.~Guestrin, ``"why should i trust you?" explaining the predictions of any classifier,'' in \emph{Proceedings of the 22nd ACM SIGKDD international conference on knowledge discovery and data mining}, 2016, pp. 1135--1144.

\bibitem{SelfExtend-Agentic-RAG}
C.~Ravuru, S.~S. Sakhinana, and V.~Runkana, ``Agentic retrieval-augmented generation for time series analysis,'' \emph{Proceedings of the 30th ACM SIGKDD Conference on Knowledge Discovery and Data Mining (KDD)}, 2024.

\bibitem{SocioDojo}
J.~Cheng and P.~Chin, ``Sociodojo: Building lifelong analytical agents with real-world text and time series,'' in \emph{International Conference on Learning Representations (ICLR)}, 2024.

\bibitem{BRIDGE}
H.~Li, Y.~Huang, C.~Xu, V.~Schlegel, R.~Jiang, R.~Batista-Navarro, G.~Nenadic, and J.~Bian, ``Bridge: Bootstrapping text to control time-series generation via multi-agent iterative optimization and diffusion modelling,'' in \emph{International Conference on Machine Learning (ICML)}, 2025.

\bibitem{TESSA}
M.~Lin, Z.~Chen, Y.~Liu, X.~Zhao, Z.~Wu, J.~Wang, X.~Zhang, S.~Wang, and H.~Chen, ``Decoding time series with llms: A multi-agent framework for cross-domain annotation,'' \emph{arXiv preprint}, 2024.

\bibitem{ma2024language}
M.~A. Merrill, M.~Tan, V.~Gupta, T.~Hartvigsen, and T.~Althoff, ``Language models still struggle to zero-shot reason about time series,'' \emph{arXiv preprint arXiv:2404.11757}, 2024.

\bibitem{TTRL}
W.~Chow, L.~E. Gardiner, H.~T. Hallgrimsson, M.~A. Xu, and S.~Y. Ren, ``Towards time-series reasoning with llms,'' in \emph{Proceedings of the NeurIPS 2024 Workshop on Time Series in the Age of Large Models}, 2024.

\bibitem{ChatTS}
Z.~Xie, Z.~Li, X.~He, L.~Xu, X.~Wen, T.~Zhang, J.~Chen, R.~Shi, and D.~Pei, ``Chatts: Aligning time series with llms via synthetic data for enhanced understanding and reasoning,'' \emph{Proceedings of the {VLDB} Endowment}, vol.~18, no.~8, pp. 2385--2398, 2025.

\bibitem{zhou2024are}
M.~Tan, M.~Merrill, V.~Gupta \emph{et~al.}, ``Are language models actually useful for time series forecasting?'' in \emph{Advances in Neural Information Processing Systems}, vol.~37, 2024, pp. 60\,162--60\,191.

\end{thebibliography}

\end{document}